\documentclass{article}

    \PassOptionsToPackage{numbers, compress}{natbib}

\usepackage{multirow}
\usepackage{multicol}
\usepackage{graphicx}
\usepackage{xcolor}
\usepackage{colortbl}
\usepackage{natbib}
\usepackage[preprint]{neurips_2025}

\usepackage[utf8]{inputenc} %
\usepackage[T1]{fontenc}    %
\usepackage{hyperref}       %
\usepackage{url}            %
\usepackage{booktabs}       %
\usepackage{amsfonts}       %
\usepackage{nicefrac}       %
\usepackage{microtype}      %
\usepackage{xcolor}         %
\usepackage{graphicx}      %
\usepackage{caption}       %
\usepackage{subcaption}    %
\usepackage{amsthm}
\usepackage{amsmath}
\usepackage{xspace}

\usepackage{colortbl}
\definecolor{baselinecolor}{gray}{.92}
\newcommand{\baseline}[1]{\cellcolor{baselinecolor}{#1}}

\newcommand{\etal}{\textit{et al.}\xspace}

\newlength\savewidth\newcommand\shline{\noalign{\global\savewidth\arrayrulewidth
  \global\arrayrulewidth 1pt}\hline\noalign{\global\arrayrulewidth\savewidth}}

\newcommand{\attname}{GRAT\xspace}
\newcommand{\attnameblock}{GRAT-B\xspace}
\newcommand{\attnamecriss}{GRAT-X\xspace}

\title{Grouping First, Attending Smartly: Training-Free Acceleration for Diffusion Transformers}

\def\eg{\emph{e.g.}} 
\def\ie{\emph{i.e.}} 
 
\def\etc{\emph{etc.}} \def\vs{\emph{vs.}}
\author{Sucheng Ren$^1$~~~ Qihang Yu$^2$~~~ Ju He$^2$~~~ Alan Yuille$^1$~~~ Liang-Chieh Chen$^2$\\
$^1$Johns Hopkins University~~~~~$^2$Independent Researcher \\
\url{https://oliverrensu.github.io/project/GRAT/}
}

\begin{document}
\maketitle

{
\begin{center}
    \centering
    \vspace{-9mm}
    \captionsetup{type=figure}
    \includegraphics[width=0.99\linewidth]{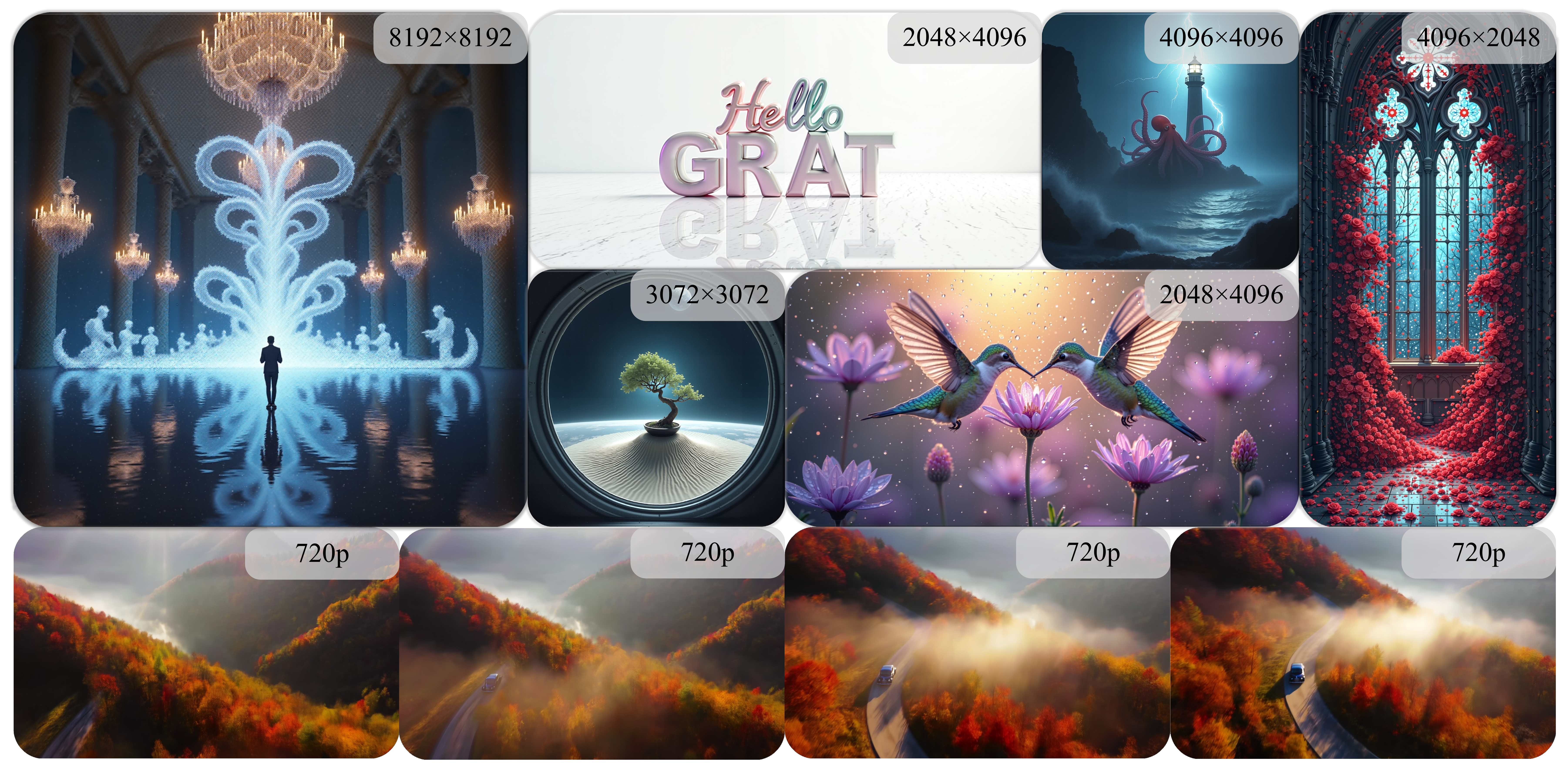}
    \captionof{figure}{
    \textbf{Fast high-resolution image and video generation enabled by equipping Flux~\cite{flux} and HunyuanVideo~\cite{hunyuan} with the proposed \attname, a training-free attention acceleration strategy.}
    \attname significantly improves inference speed without any fine-tuning or degradation in visual quality.
    }
    \label{fig:teaser}
\end{center}%
}
\begin{abstract}
Diffusion-based Transformers have demonstrated impressive generative capabilities, but their high computational costs hinder practical deployment—for example, generating an $8192\times8192$ image can take over an hour on an A100 GPU.
In this work, we propose \attname (\textbf{GR}ouping first, \textbf{AT}tending smartly), a training-free attention acceleration strategy for fast image and video generation without compromising output quality.
The key insight is to exploit the inherent sparsity in learned attention maps (which tend to be locally focused) in pretrained Diffusion Transformers and leverage better GPU parallelism.
Specifically, \attname first partitions contiguous tokens into non-overlapping groups, aligning both with GPU execution patterns and the local attention structures learned in pretrained generative Transformers.
It then accelerates attention by having all query tokens within the same group share a common set of attendable key and value tokens. These key and value tokens are further restricted to structured regions, such as surrounding blocks or criss-cross regions, significantly reducing computational overhead (\eg, attaining a \textbf{35.8$\times$} speedup over full attention when generating $8192\times8192$ images) while preserving essential attention patterns and long-range context.
We validate \attname on pretrained Flux and HunyuanVideo for image and video generation, respectively.
In both cases, \attname achieves substantially faster inference without any fine-tuning, while maintaining the performance of full attention.
We hope \attname will inspire future research on accelerating Diffusion Transformers for scalable visual generation.
Code is available at \url{https://github.com/OliverRensu/GRAT}.
\end{abstract}

\section{Introduction}
Diffusion and flow-matching generative models~\cite{diff1,diff2,sd,lipmanflow,liuflow}, when paired with Transformer backbones~\cite{vaswani2017attention,dit,bao2023all,sit,liu2024alleviating,shin2025deeply}, have recently achieved state-of-the-art performance across a wide range of image and video generation tasks~\cite{blattmann2023stable,sd3,flux,hunyuan,wan,yu2024randomized,ren2025flowar,ren2025beyond,deng2025coconut,he2025flowtok,liu2025revision}.
Despite their success, practical deployment remains challenging due to the prohibitive $\mathcal{O}(N^2)$ computational cost of self-attention, which scales quadratically with sequence length. This computational overhead significantly increases inference latency and resource consumption, ultimately hindering the adoption of Diffusion Transformers~\cite{dit,sit} in real-world creative and interactive applications.

Reducing inference cost requires optimizing the attention mechanism~\cite{vaswani2017attention}, which remains the dominant computational bottleneck. Although self-attention is designed as a global operation, most visual dependencies in image and video data are inherently sparse due to significant spatial redundancy~\cite{he2022masked,yu2024image,kim2025democratizing}. This raises a fundamental question:
\emph{Is it necessary to retain full global attention for every token?}

\begin{figure}
    \centering
  \begin{subfigure}{0.31\textwidth}
    \includegraphics[width=\linewidth]{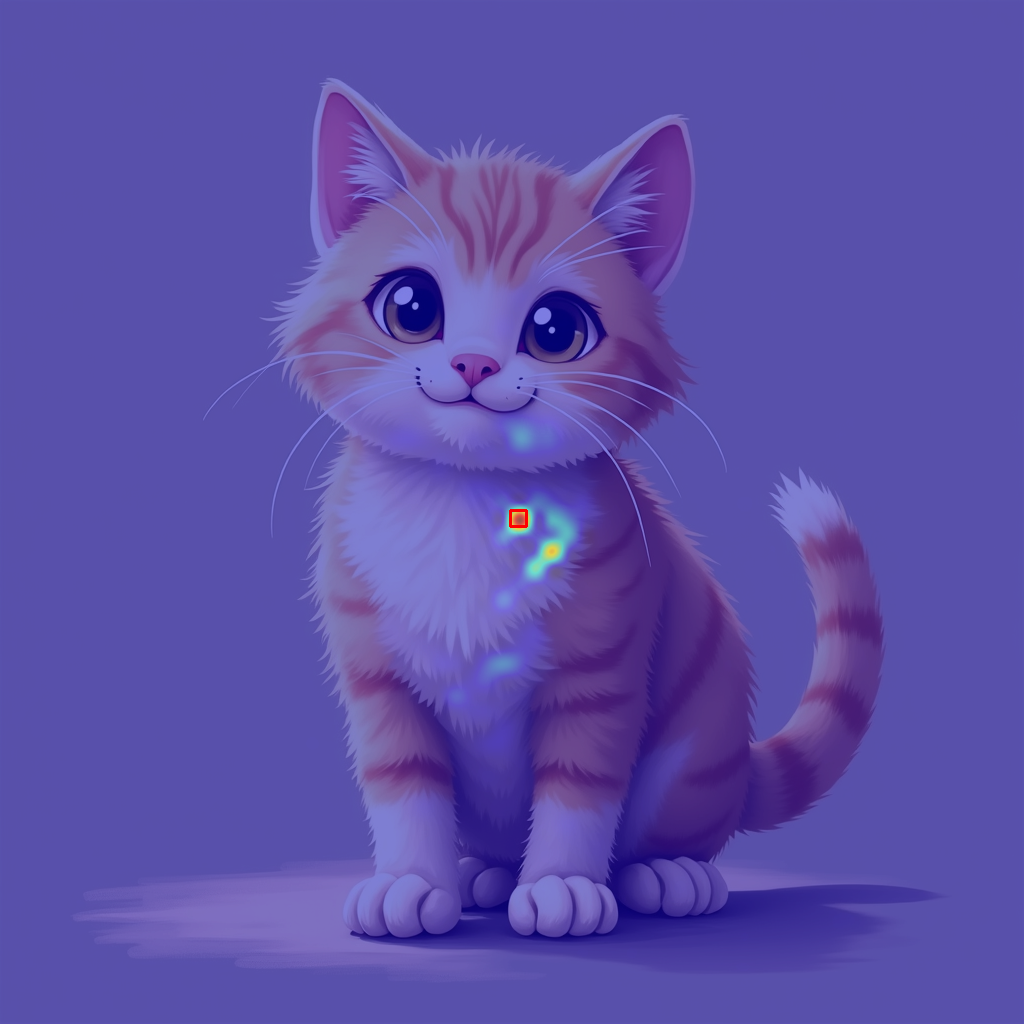}
  \end{subfigure}
  \hfill
  \begin{subfigure}{0.31\textwidth}
    \includegraphics[width=\linewidth]{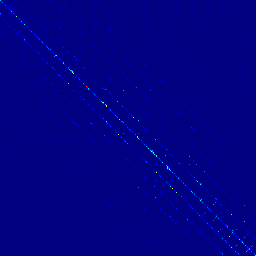}
  \end{subfigure}
  \hfill
  \begin{subfigure}{0.35\textwidth}
    \raisebox{-5mm}{%
      \includegraphics[width=\linewidth]{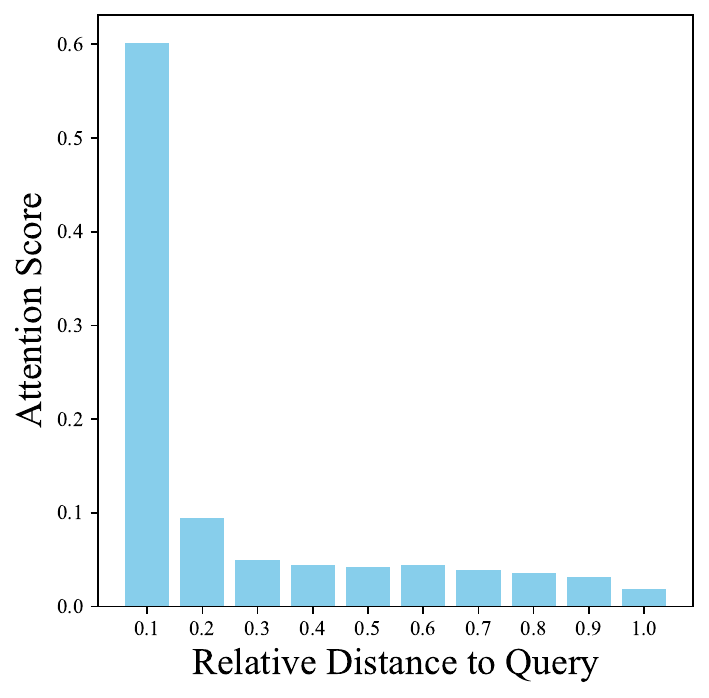}
    }
  \end{subfigure}
  \vspace{-1mm}
    \caption{
    \textbf{Attention Visualization of Flux~\cite{flux}.}
    \textbf{\emph{Left:}} A single query token (marked in red) attends only to sparse, local regions.
    \textbf{\emph{Middle:}} Visualization of attention maps across all query tokens in an image (each row corresponds to a query). The patterns consistently demonstrate sparsity and locality.
    \textbf{\emph{Right:}} Distribution of attention scores, averaged over 100 generated images, plotted as a function of the normalized spatial distance between query and key tokens (1.0 denotes the maximum distance).
    The attention score—defined as the scaled dot product between query and key tokens after softmax normalization—reflects the contribution of each query-key pair to the final output.
    The distribution is long-tailed, indicating that spatially close key tokens contribute most to the attention output.
    }
    \label{fig:visual_attn}
\end{figure}

We investigate this question by visualizing attention maps from the modern text-to-image model Flux~\cite{flux} in Fig.\ref{fig:visual_attn}.
Specifically, Fig.\ref{fig:visual_attn} (Left) highlights that the attention of a given query token (marked in red) is highly sparse and localized around its spatial neighborhood.
Fig.~\ref{fig:visual_attn} (Middle) shows attention maps across all query tokens (each row corresponding to a query), revealing that this sparsity and locality are consistent throughout the image.
To further quantify attention patterns, we average attention scores across 100 generated images and plot them as a function of normalized spatial distance between query and key tokens (where 1.0 denotes the maximum possible distance). The attention score—defined as the scaled dot product between a query and key after softmax normalization—determines how much each value contributes to the final output.
As shown in Fig.~\ref{fig:visual_attn} (Right), the distribution exhibits a long-tailed pattern, where spatially close tokens dominate. Notably, key tokens within a 0.2 normalized distance contribute roughly 69\% of the total attention mass.
These findings confirm that attention in pretrained Diffusion Transformers~\cite{flux} is predominantly sparse and local, with long-range interactions needed for only a small subset of tokens. As a result, much of the computation performed by full self-attention is redundant, leading to significant inefficiency and unnecessary GPU usage in modern Diffusion Transformer models~\cite{flux,hunyuan}.

Prior works have sought to address this inefficiency by limiting each query's attention to a local region~\cite{ramachandran2019stand,wang2020axial,na,he2023maxtron}, reducing the theoretical complexity from quadratic to linear~\cite{clear,sta}.
For instance, CLEAR~\cite{clear} reports reduced FLOPs by restricting each query's receptive field to a fixed-radius  neighborhood. However, we identify three practical drawbacks that limit such approaches for fast Diffusion Transformers:
(1) Limited receptive field — keeping the radius small preserves efficiency but refrains the model from capturing long-range dependencies critical for global coherence;
(2) Quality degradation — performance drops without additional fine-tuning, which is costly and undermines plug-and-play deployment;
(3) Poor hardware utilization — since each query token attends to a different set of key-value tokens (via sliding circular windows), attention computation becomes irregular and inefficient on GPUs.
These limitations often negate the theoretical benefits of reduced FLOPs, resulting in little real-world inference speedup.

\begin{figure}
    \centering
    \includegraphics[width=0.85\linewidth]{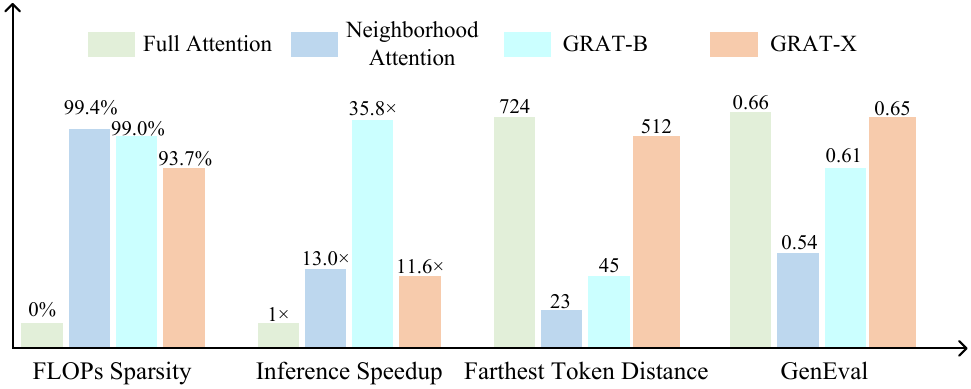}
    \vspace{-1mm}
    \caption{
    \textbf{Comparison of Attention Schemes.}
    The comparison is based on Flux~\cite{flux} with various attention mechanisms, including Full Attention~\cite{vaswani2017attention}, Neighborhood Attention~\cite{ramachandran2019stand,na}, and the proposed \attnameblock and \attnamecriss.
    \emph{FLOPs Sparsity} measures the theoretical reduction in compute relative to Full Attention (0\% indicates no reduction).
    \emph{Inference Speedup} reflects real-world speedup on an A100 GPU, relative to Full Attention (1$\times$ means no speedup).
    \emph{Farthest Token Distance} denotes the maximum distance over which a query can attend—representing the effective receptive field.
    As shown, \attnameblock achieves the same FLOPs sparsity as Neighborhood Attention (NA) but delivers higher inference speedup. Conversely, \attnamecriss maintains comparable speedup while offering a much larger receptive field. Both variants outperform NA in GenEval scores~\cite{geneval}, with \attnamecriss notably matching the quality of Full Attention while running 12$\times$ faster.
    }
    \label{fig:performance}
\end{figure}

To overcome these challenges, we propose \attname (\emph{\textbf{GR}ouping first, \textbf{AT}tending smartly}), a training-free attention acceleration strategy that preserves the generation quality of pretrained Diffusion Transformers without any fine-tuning, while delivering substantial improvements in inference efficiency.
\attname operates in two stages. The first stage, \emph{Grouping First}, partitions the input feature map into non-overlapping square groups. Group sizes are selected to align with GPU streaming multiprocessor (SM) and thread block configurations, maximizing hardware parallelism. This layout not only improves GPU utilization but also reflects the empirical bias that diffusion attention is predominantly local.
In the second stage, \emph{Attending Smartly}, each query group restricts its attention to a structured region rather than the entire feature map. This design is motivated by two empirical observations:
(1) the majority of attention mass is concentrated in a query token’s local spatial neighborhood (as shown in Fig.~\ref{fig:visual_attn}); and
(2) preserving a small set of long-range dependencies is sufficient to maintain large receptive fields and global coherence.
Based on these insights, we introduce two variants: \attnameblock, where each query group attends to its surrounding blocks, and \attnamecriss, which enables long-range attention by attending across rows and columns.

When integrated into the pretrained Diffusion Transformer Flux~\cite{flux}, \attname significantly accelerates the attention mechanism.
As shown in Fig.~\ref{fig:performance}, under comparable FLOPs sparsity, \attnameblock is approximately 2.85$\times$ faster than Neighborhood Attention (NA)~\cite{na}, while attending to 1.96$\times$ more distant tokens.
At a similar inference speed to NA, \attnamecriss preserves more long-range context, attending to tokens up to 22$\times$ farther.
We further apply \attname to the modern video generation model HunyuanVideo~\cite{hunyuan} and observe similarly strong results.
\attnameblock achieves a 15.8$\times$ speedup over full attention while outperforming prior work~\cite{sta} in both speed and sample quality.
Meanwhile, \attnamecriss matches the performance of full attention while running 2.4$\times$ faster.

\section{Related Work}
Most prior efforts to accelerate diffusion models have focused on model-level optimizations—such as model compression~\citep{kim2024architecturalcompressiontexttoimagediffusion,lee2024koalaselfattentionmattersknowledge,fang2023structuralpruningdiffusionmodels,shang2023posttrainingquantizationdiffusionmodels,li2023qdiffusionquantizingdiffusionmodels,yang20241} and reducing the number of inference steps~\citep{diff1,liu2022pseudonumericalmethodsdiffusion,salimans2022progressivedistillationfastsampling}—as well as attention-centric acceleration techniques. In this section, we focus on the latter.

\textbf{Compressed Key and Value.}
The self-attention mechanism in Transformers~\cite{vaswani2017attention} has quadratic complexity in the number of tokens, which becomes a major bottleneck for high-resolution diffusion models. One line of research reduces this cost by compressing the key and value sequences.
Linformer~\cite{linformer} projects the length-$N$ key and value vectors into a lower-dimensional subspace of size $k \ll N$, reducing attention complexity to $\mathcal{O}(Nk)$.
PixArt-$\Sigma$~\cite{pixart} compresses key and value representations using a 2D strided convolution (stride $s$), which effectively reduces the sequence length by a factor of $s^2$.
Other approaches leverage learned pooling~\cite{lee2019settransformerframeworkattentionbased} or token clustering/pruning strategies~\cite{Wu2023PPT} to produce compact key/value sets.
Recent work has applied such ideas to diffusion models, such as grouped or windowed attention in Diffusion Transformers~\cite{yuan2024ditfastattnattentioncompressiondiffusion}, to reduce attention overhead.
While these methods improve efficiency, they often introduce aliasing or mix fine-grained details with background context—leading to degradation in generation quality.

\textbf{Sparse Key and Value.}
An alternative to compression is to enforce \emph{sparse} attention, where each query attends only to a limited subset of key-value tokens.
This approach is motivated by empirical findings that high-magnitude attention weights are typically concentrated in local regions~\cite{xie2023revealing}.
In NLP, Longformer~\cite{zaheer2021bigbirdtransformerslonger} achieves linear complexity by combining sliding-window attention with a few global tokens.
Vision Transformer~\cite{dosovitskiy2020image} variants adopt a similar principle: Swin Transformer~\cite{swin} uses non-overlapping windows with periodic shifts to enable cross-window interaction.
Stand-Alone Self-Attention~\cite{ramachandran2019stand} and Neighborhood Attention~\cite{na} similarly restrict attention to spatially local neighborhoods, achieving linear complexity in time and memory.

Within Diffusion Transformer models~\cite{dit}, CLEAR~\cite{clear} restricts attention to circular local windows.
Zhang \etal extend this idea to the temporal domain using a sliding-tile attention (STA) mechanism~\cite{sta}, which confines attention to local spatial-temporal tiles.
However, STA’s relatively large windows reduce overall sparsity, limiting its maximum achievable speedup. Its inability to capture long-range dependencies further degrades generation quality.
In contrast, our proposed \attname groups tokens and constrains the attention computation to structured regions, enabling both higher sparsity and improved GPU parallelism.
This design leads to significantly faster inference while preserving sample quality, outperforming prior sparse attention methods that rely on overly restrictive locality.

\section{Method}

\begin{figure*}[t]
    \centering
    \includegraphics[width=\linewidth]{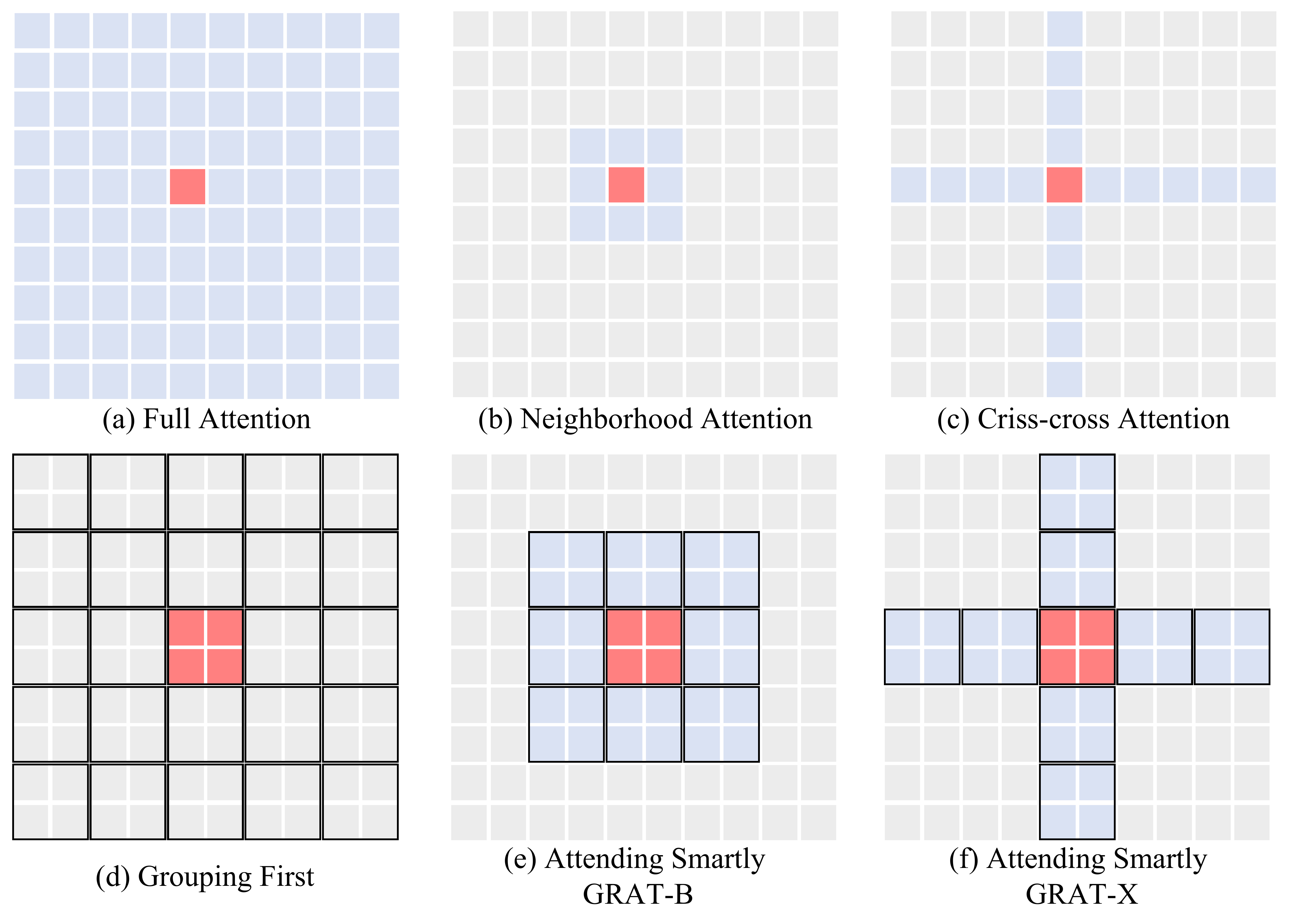}
    \caption{
    \textbf{Illustration of Attention Operations.}
    Query tokens are shown in red, and their corresponding attended regions (key and value tokens) are highlighted in light blue.
    (a) Full Attention~\cite{vaswani2017attention}: each query attends to the entire feature map.
    (b) Neighborhood Attention~\cite{ramachandran2019stand,na}: each query attends only to its local spatial neighborhood.
    (c) Criss-cross Attention: each query attends to tokens in the same row and column.
    (d–f) In this work, we propose \attname (\textbf{GR}ouping first, \textbf{AT}tending smartly), which first partitions the feature map into non-overlapping groups (d) (each group has $2 \times 2$ tokens in this example). Query tokens within the same group share a common set of key and value tokens, which are restricted to structured regions—such as surrounding blocks (e) or criss-cross patterns (f).
    }
    \label{fig:method}
\end{figure*}

Our goal is to accelerate inference in pretrained Diffusion Transformers (DiTs)~\cite{dit,sit,flux,hunyuan} \emph{without any additional fine-tuning}, while preserving sample quality.
In this section, we first revisit the computational bottlenecks of full self-attention~\cite{vaswani2017attention} (Sec.~\ref{sec:prelim}), and then introduce our attention acceleration strategy: \emph{Grouping First, Attending Smartly} (Sec.~\ref{sec:our_method}).

\subsection{Preliminary: Self-Attention}
\label{sec:prelim}
Given a token sequence $\mathbf{X}\!=\![x_{1,1},\dots,x_{H,W}]\!\in\!\mathbb{R}^{N\times d}$, where $N = H \times W$ denotes the number of tokens (with $H$ and $W$ being the height and width of the feature map), and $d$ denotes the feature dimension, the standard self-attention~\cite{vaswani2017attention} is computed as follows:
\begin{equation}
    \mathrm{Attn}(\mathbf{Q},\mathbf{K},\mathbf{V}) =
\mathrm{softmax}(\frac{\mathbf{Q}\mathbf{K}^{\top}}{\sqrt{d}}\bigr)\mathbf{V},
\end{equation}
where the query, key, and value tokens are obtained by linearly projecting the input tokens: $\mathbf{Q}\!=\!\mathbf{X}\mathbf{W}_{Q}$, $\mathbf{K}\!=\!\mathbf{X}\mathbf{W}_{K}$, and $\mathbf{V}\!=\!\mathbf{X}\mathbf{W}_{V}$.  
This formulation incurs a quadratic computational cost of $\mathcal{O}(N^2)$, which quickly becomes the dominant bottleneck—especially for high-resolution images or long video sequences. For instance, generating an $8192\times8192$ image, even with a VAE~\cite{vae} that downsamples the spatial resolution by a factor of 16, still requires $N = 262{,}144$ tokens, making full attention prohibitively expensive.
This challenge is further exacerbated in models based on Diffusion Transformers~\cite{dit}, such as Flux~\cite{flux} and HunyuanVideo~\cite{hunyuan}, where the attention modules are invoked repeatedly along the denoising trajectory.

To address this, we propose reducing the effective sequence length that each query token attends to—transforming the quadratic complexity into near-linear complexity—while achieving real-world speedups by leveraging the parallelism-friendly layout of GPUs.

\subsection{\attname: A Training-Free Attention Acceleration Strategy}
\label{sec:our_method}
To accelerate the attention operation, we propose \attname, a training-free attention acceleration strategy (Fig.~\ref{fig:method}).
It consists of two key steps: (1) \emph{grouping first}, which partitions contiguous tokens into non-overlapping groups, and (2) \emph{attending smartly}, which restricts the attended keys and values to structured regions.
As a result, \attname significantly reduces computational overhead while preserving essential attention patterns and long-range context.
We describe each step in detail below.

\textbf{Grouping First.}
Given the query, key, value tokens (\ie, $\mathbf{Q}$, $\mathbf{K}$, and $\mathbf{V}$), \attname takes their spatial positions into account and partitions them into non-overlapping groups, where each group contains $g_1 \times g_2$ tokens. The grouping is defined as follows:
\begin{align}
    \mathcal{G}^{Q}_{p,q} &= \bigl\{Q_{i,j} \bigm| i//g_1=p, j// g_2 = q\bigr\}, \\
    \mathcal{G}^{K}_{p,q} &= \bigl\{K_{i,j} \bigm| i//g_1=p, j// g_2 = q\bigr\}, \\
    \mathcal{G}^{V}_{p,q} &= \bigl\{V_{i,j} \bigm| i//g_1=p, j// g_2 = q\bigr\}, 
\end{align}
where $i$ and $j$ index the token’s spatial position, and $p$ and $q$ index the group. Tokens within the same group $\mathcal{G}_{p,q}$ are contiguous in memory. Superscripts denote the token type (query, key, or value).

This design yields two key benefits:
(1) coalesced memory access that improves GPU throughput, and
(2) an inductive bias aligned with the local attention patterns commonly observed in pretrained Diffusion Transformer models.

\textbf{Attending Smartly.}
Motivated by the observation that learned attention maps in Diffusion Transformers~\cite{flux} exhibit strong sparsity, \attname restricts the attendable key and value tokens to structured regions.
As a result, attention is computed only over these sparse, structured regions, significantly reducing computational cost.

In this work, we explore two types of structured regions: (1) surrounding blocks (\attnameblock), and (2) criss-cross regions (\attnamecriss).
For surrounding blocks, each query group $\mathcal{G}^Q_{p,q}$ attends only to keys and values within its $(2b+1) \times (2b+1)$ neighboring groups. The attendable sets are defined as: 
\begin{align}
\mathcal{G}'^{K}_{p,q}
&= \bigcup_{\{(m,n) \ \mid \ |m-p|\le b \;\land\; |n-q|\le b\}}
   \mathcal{G}^{K}_{m,n},
\\
\mathcal{G}'^{V}_{p,q}
&= \bigcup_{\{(m,n) \ \mid \ |m-p|\le b \;\land\; |n-q|\le b\}}
   \mathcal{G}^{V}_{m,n},
\end{align}
where $\mathcal{G}'^{K}_{p,q}$ $\mathcal{G}'^{V}_{p,q}$ denote the attendable key and value groups for the query group $\mathcal{G}^{Q}_{p,q}$, and $b$ is the surrounding block size (we use $b=1$ by default).

This design reduces the attention complexity from $\mathcal{O}(N^2)$ to $\mathcal{O}(b^2 g^2 N)$, which becomes linear in $N$ when $g$ and $b$ are constants.
Since $g \ll N$ in practice, \attnameblock achieves substantial computational savings and real-world inference speedups.

The ``surrounding blocks'' design restricts the receptive field to nearby groups, limiting its ability to capture long-range dependencies.
To address this, we introduce ``criss-cross regions'', which incorporate long-range attention by allowing each query group to attend along its entire row and column, resulting in the \attnamecriss variant.
Specifically, for each query group $\mathcal{G}^Q_{p,q}$, the attendable key and value groups are defined as:
\begin{align}
\mathcal{G}'^{K}_{p,q}
&= \bigcup_{\{(m,n) \ \mid \ m = p \;\lor\; n=p\}}
   \mathcal{G}^{K}_{m,n},
\\
\mathcal{G}'^{V}_{p,q}
&= \bigcup_{\{(m,n) \ \mid \ m=p \;\lor\; n=p\}}
   \mathcal{G}^{V}_{m,n}.
\end{align}
This design yields an attention complexity of $\mathcal{O}(gHN + gWN)$.
Although \attnamecriss incurs higher computational overhead than \attnameblock, it offers better performance by incorporating long-range contextual information across both row and column dimensions.

\textbf{Extension to Videos.}
Our \attname\ framework generalizes naturally to video by treating time as an additional dimension in the ``Grouping First'' step. 
Given a video feature map of shape $T \times H \times W$ (\ie, a video of $T$ frames), we partition the feature map into non‐overlapping 3D groups of size $g_1 \times g_2 \times g_3$, where $g_1$ indexes the temporal frames and $g_2$ and $g_3$ index spatial height and width, respectively.
Specifically, each group is defined as:
\begin{align}
    \mathcal{G}^{Q}_{p_1,p_2,p_3} &= \bigl\{Q_{t,i,j} \bigm| t//g_1=p_1, i//g_2=p_2, j// g_3 = p_3\bigr\}, \\
    \mathcal{G}^{K}_{p_1,p_2,p_3} &= \bigl\{K_{t,i,j} \bigm| t//g_1=p_1, i//g_2=p_2, j// g_3 = p_3\bigr\}, \\
    \mathcal{G}^{V}_{p_1,p_2,p_3} &= \bigl\{V_{t,i,j} \bigm| t//g_1=p_1, i//g_2=p_2, j// g_3 = p_3\bigr\}. 
\end{align}
In the second ``Attending Smartly'' step, we generalize the structured regions to their spatio-temporal counterparts: namely, \emph{3D Surrounding Blocks} and \emph{3D Criss-Cross Regions}.
For \attnameblock, each query group $\mathcal{G}^{Q}_{p_1,p_2,p_3}$ attends only to the $(2b_t+1)\times(2b_h+1)\times(2b_w+1)$ neighboring groups:
\begin{align}
\mathcal{G}'^{K}_{p'_1,p'_2,p'_3}
& = \bigcup_{\{(p_1,p_2,p_3) \ \mid \ |p_1-p'_1|\le b_t \;\land\; |p_2-p'_2|\le b_h\;\land\; |p_3-p'_3|\le b_w\}}
   \mathcal{G}^{K}_{p_1,p_2,p_3},
   \\
\mathcal{G}'^{V}_{p'_1,p'_2,p'_3}
& = \bigcup_{\{(p_1,p_2,p_3) \ \mid \ |p_1-p'_1|\le b_t \;\land\; |p_2-p'_2|\le b_h\;\land\; |p_3-p'_3|\le b_w\}}
   \mathcal{G}^{V}_{p_1,p_2,p_3}.
\end{align}

For \attnamecriss, each query group attends along its entire temporal, vertical, and horizontal axes to capture long-range dependencies:
\begin{align}
\mathcal{G}'^{K}_{p'_1,p'_2,p'_3}
& = \bigcup_{\{(p_1,p_2,p_3) \ \mid \ p_1=p'_1 \;\lor\; \ p_2=p'_2  \;\lor\; \ p_3=p'_3\}}
   \mathcal{G}^{K}_{p_1,p_2,p_3},
   \\
\mathcal{G}'^{V}_{p'_1,p'_2,p'_3}
& = \bigcup_{\{(p_1,p_2,p_3) \ \mid \ p_1=p'_1 \;\lor\; \ p_2=p'_2  \;\lor\; \ p_3=p'_3\}}
   \mathcal{G}^{V}_{p_1,p_2,p_3}.
\end{align}

\section{Experimental Results}
We propose \attname, a training-free attention acceleration strategy, and validate its effectiveness on two state-of-the-art pretrained models.
We first evaluate it on Flux~\cite{flux} for text-to-image generation in Sec.~\ref{sec:result_flux}), and then on HunyuanVideo~\cite{hunyuan} for text-to-video generation in Sec.~\ref{sec:result_hunyuan}.
% Due to space constraints, the ablation study is provided in the Appendix.
The ablation study is provided in the Appendix.

\begin{table}[t]
    \centering
    \small
    \caption{
    \textbf{Forward Speed of Sparse Attentions for Image Generation.}
    All methods are implemented on top of Flux.1-dev~\cite{flux} to generate $8192 \times 8192$ images, resulting in a sequence length of 262,144 tokens.
    Evaluation is conducted on a single A100 GPU.
    \emph{FLOPs sparsity} indicates the theoretical reduction in compute relative to Full Attention (0\% means no reduction).
    \emph{Farthest token} denotes the maximum distance a query can attend to (724 is the maximum in this setting).
    \emph{Attention latency} measures the time required for the attention operation alone.
    \emph{Inference speedup} reflects the acceleration of the attention module relative to Full Attention (1$\times$ means no speedup).
    \emph{Inference time} captures the total runtime of the full model during inference.
    }
    \label{tab:speed_image}
    \scalebox{0.96}{
    \begin{tabular}{c|c|ccccc} 
      \multirow{2}{*}{method}   & \multirow{2}{*}{configuration} & FLOPs & farthest & attention& inference  & inference time  \\
      & & sparsity & token& latency (s) & speedup &(sec/image)\\
      \shline
      Full Attention~\cite{flux,vaswani2017attention,flashattention2}  &N/A &0\%&724 & 4.081& 1.0$\times$&5480 \\
      CLEAR~\cite{clear} &radius=16&99.50\% &16 &0.280&14.6$\times$ &812 \\
      NA~\cite{na,ramachandran2019stand} &window=32&99.42\% & 23&0.312 &13.0$\times$ & 842\\
      \hline
      \attnameblock &b=1&99.03\%& 45&0.114&35.8$\times$&598 \\
      \attnamecriss &N/A &93.67\% & 512&0.353&11.6$\times$&898 \\
    \end{tabular}
    }
\end{table}

\subsection{Text-to-Image Generation with Flux}
\label{sec:result_flux}

\textbf{Experimental Setup.}
We conduct experiments using the state-of-the-art Flux model~\cite{flux}, with full attention implemented via FlashAttention-2~\cite{flashattention2}.
All sparse attention baselines—including our proposed method—are implemented using Flex Attention~\cite{flex} within the PyTorch framework~\cite{paszke2019pytorch}.
For all \attname variants, we adopt a group size of $16 \times 16$.
Following standard evaluation protocols, we report Fréchet Inception Distance (FID)~\cite{fid}, Image Reward (IR)~\cite{imagereward}, and CLIP text–image similarity (CLIP-T)~\cite{clip} on the COCO2014~\cite{coco} and MJHQ-30K~\cite{li2024playground} datasets.
We additionally evaluate on GenEval~\cite{geneval}, an object-centric benchmark designed to assess fine-grained generation quality.

\textbf{Efficiency.}
Tab.~\ref{tab:speed_image} summarizes the efficiency comparison across various attention mechanisms.
Our \attnameblock achieves the highest inference speedup. Compared to the full attention baseline, it reduces attention latency from 4.081 seconds to 0.114 seconds—a \textbf{35.8$\times$} speedup—and shortens end-to-end generation time by \textbf{9.2$\times$}.
Its FLOPs sparsity (99.03\%) is comparable to that of CLEAR ($r=16$)~\cite{clear} and Neighborhood Attention (NA) with $w=32$~\cite{na}, yet \attnameblock still outperforms them due to more efficient GPU parallelism.
In addition, \attnamecriss achieves a similar inference speed to NA ($w=32$)—with attention latency of 0.354 \vs \ 0.312 (sec) and overall generation time of 898 \vs \ 842 (sec/image)—while retaining significantly more attention capacity.
Notably, it supports a farthest-token distance of 512, compared to just 23 in NA—\textbf{22$\times$} farther—demonstrating a much larger effective receptive field at a similar runtime cost.

\begin{table}[t]
\centering
\caption{
\textbf{Quantitative Results of Different Attention Mechanisms on COCO2014 Validation Set~\cite{coco}, MJHQ-30K~\cite{li2024playground}, and GenEval~\cite{geneval}}
All methods use Flux.1-dev~\cite{flux}.
We report FID~\cite{fid}, Image Reward (IR)~\cite{imagereward}, CLIP text–image similarity (CLIP-T)~\cite{clip}, and GenEval overall scores.
}
\begin{tabular}{c|ccc|ccc|c}
\multirow{2}{*}{method} & \multicolumn{3}{c|}{COCO2014} & \multicolumn{3}{c|}{MJHQ-30K} & GenEval\\
                       & FID$\downarrow$    & IR$\uparrow$ &CLIP-T$\uparrow$   & FID$\downarrow$    & IR $\uparrow$ &CLIP-T$\uparrow$  
                       & overall \\
\shline
  Full Attention~\cite{flux,vaswani2017attention}            &   33.89  &        1.076         &    26.03    &   19.72  &0.938  &26.30 & 0.66  \\
  CLEAR~\cite{clear}       &     47.50   &         0.045        &      24.79  &  29.96   &0.270 &25.60  &0.52  \\
  NA~\cite{na} &42.62&0.112&24.80&28.65&0.333&25.82 & 0.54\\
\hline
\attnameblock  & 35.99&         0.925      & 25.83 &20.95& 0.900&26.26 & 0.61\\
\attnamecriss &  34.59&           1.068   & 26.17 &20.05 &0.998&26.32 &0.65\\
\end{tabular}
\label{tab:coco}
\end{table}

\begin{figure}[!t]
    \centering
    \setlength{\tabcolsep}{2 pt}
    \begin{tabular}{ccccc}
    \includegraphics[width=0.19\linewidth]{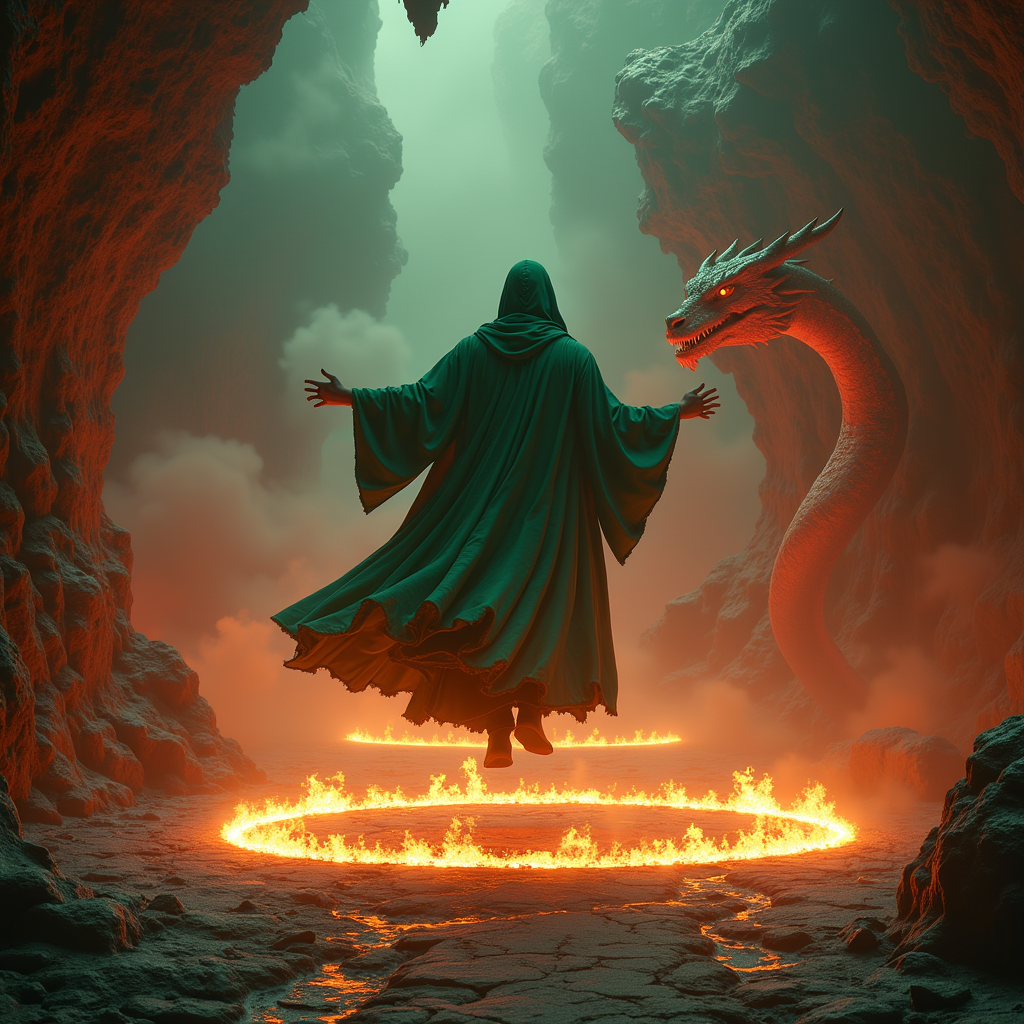} & 
        \includegraphics[width=0.19\linewidth]{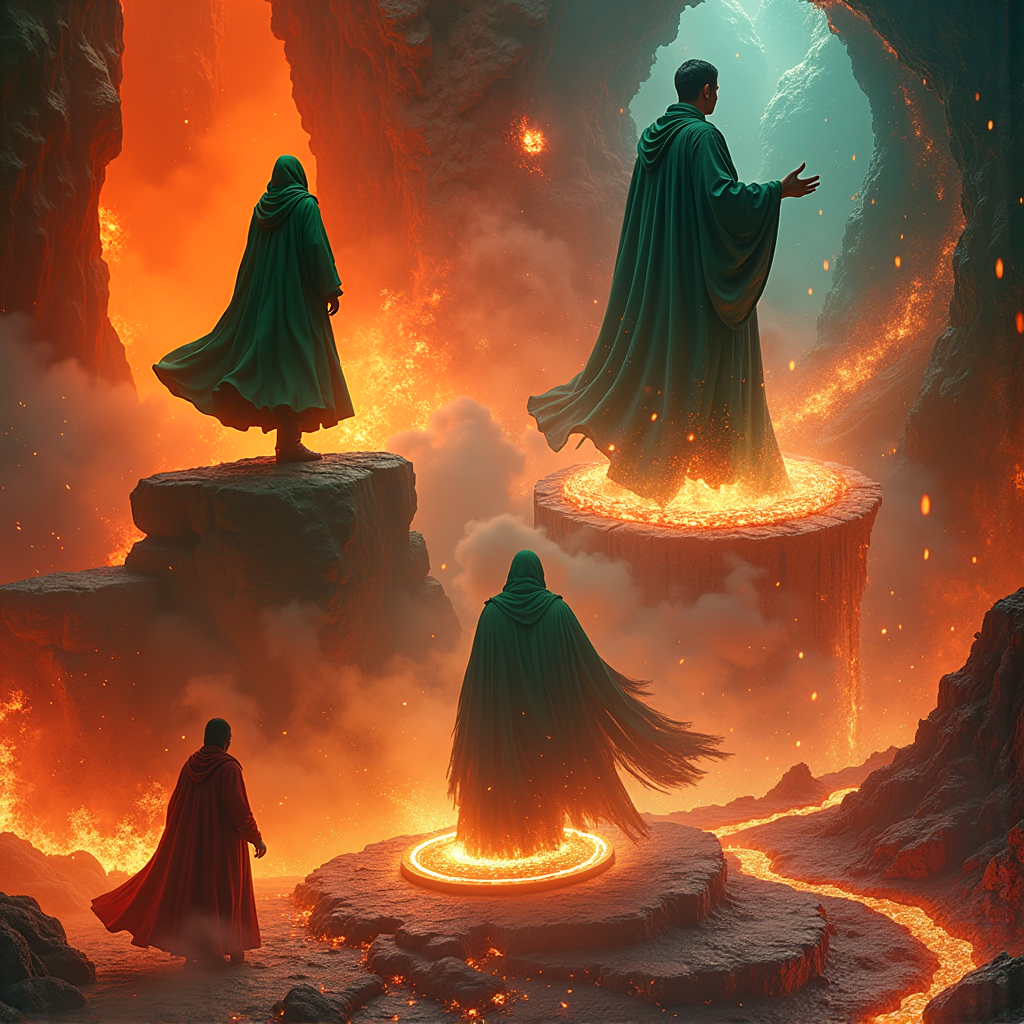} & \includegraphics[width=0.19\linewidth]{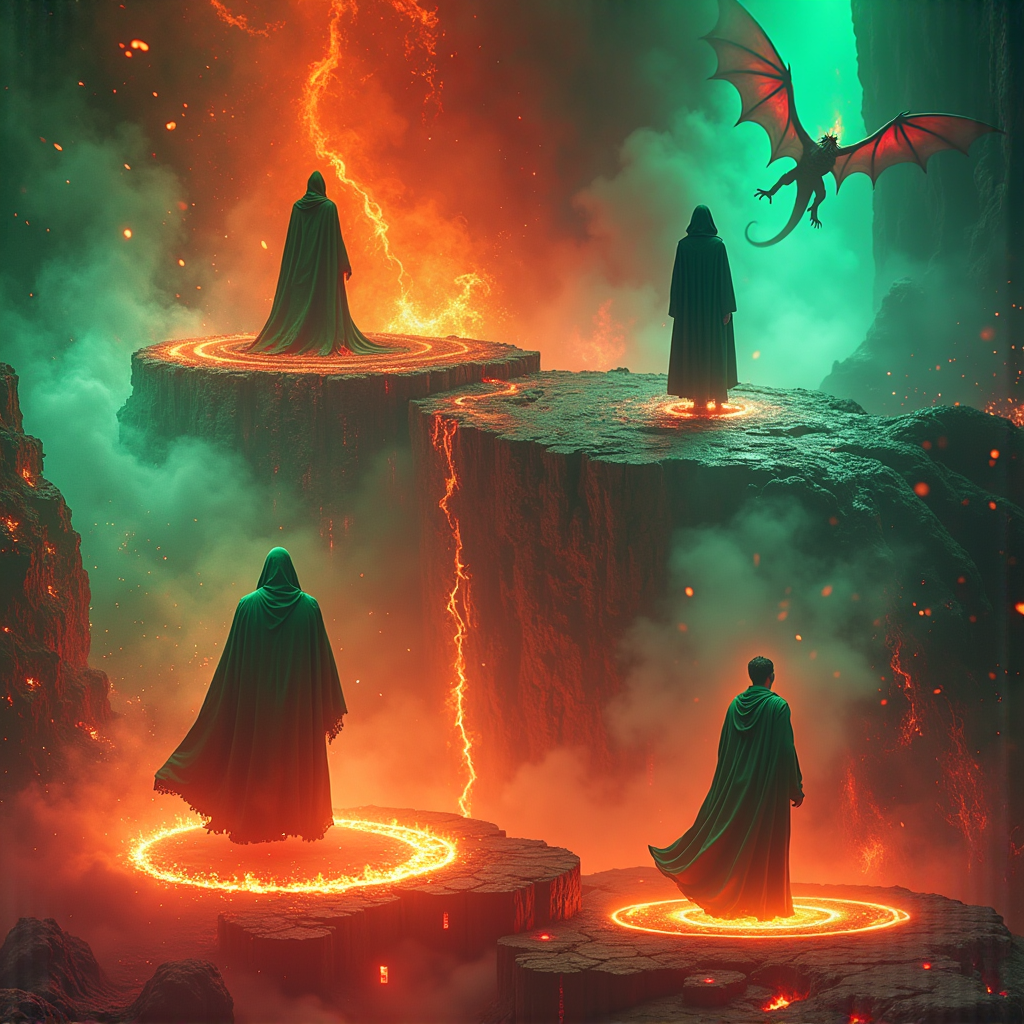} & \includegraphics[width=0.19\linewidth]{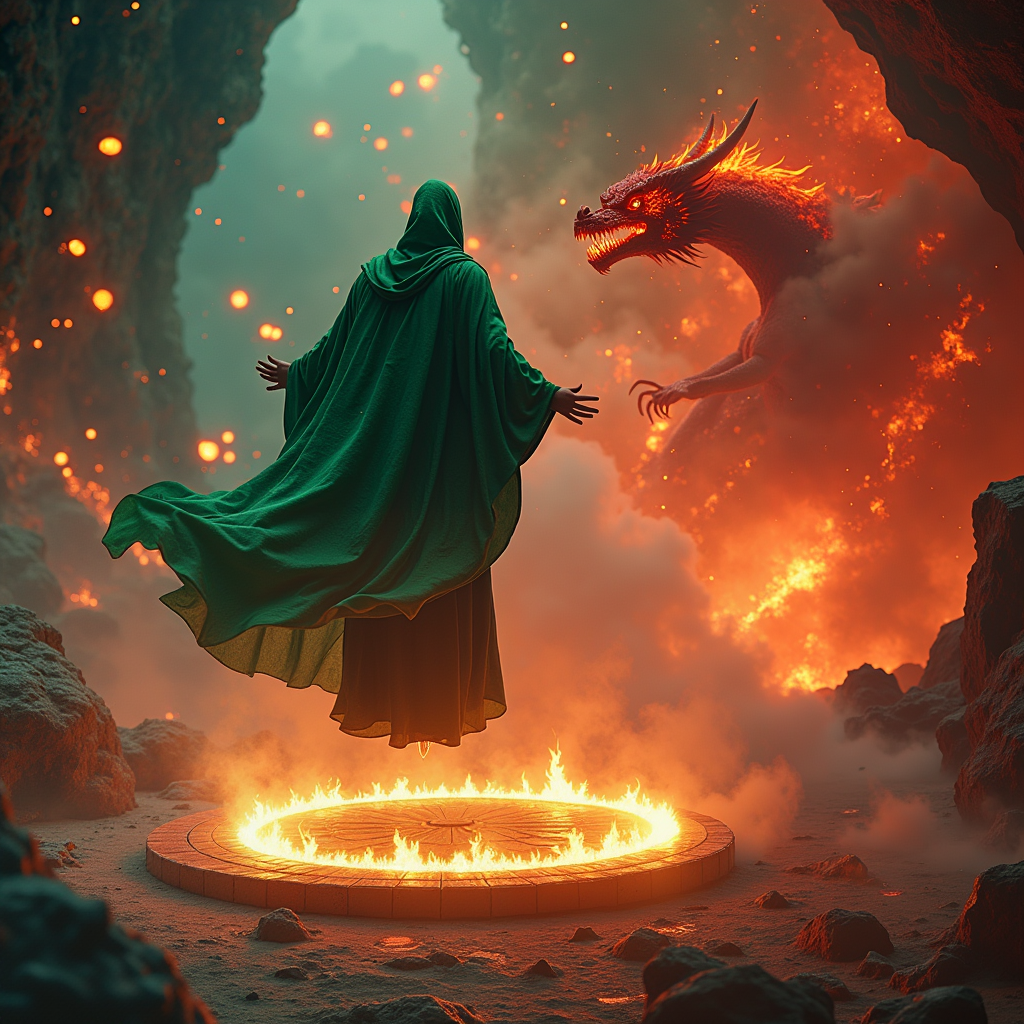} & \includegraphics[width=0.19\linewidth]{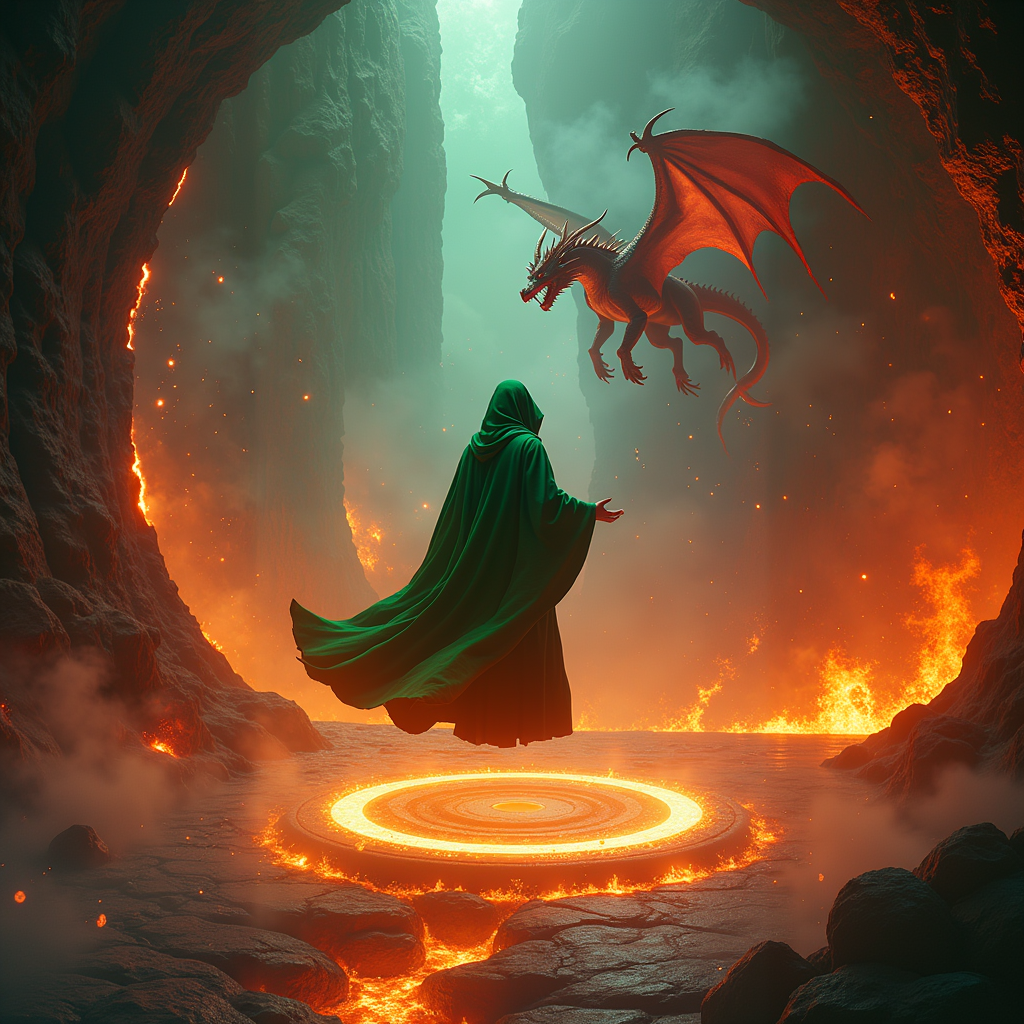}  \\
        Full Attention&NA & CLEAR& \attnameblock &\attnamecriss\\
    \end{tabular}
    \vspace{-2mm}
    \caption{
    \textbf{Generated Images by Flux~\cite{flux} with Different Attention Mechanisms.}
    We compare the visual results of Full Attention~\cite{vaswani2017attention} (\ie, the original Flux), Neighborhood Attention (NA)~\cite{na}, CLEAR~\cite{clear}, and our proposed \attnameblock and \attnamecriss.
    }
    \label{fig:visual_image}
\end{figure}
\vspace{-2mm}

\textbf{Quantitative Comparison.}
Tab.~\ref{tab:coco} presents quantitative results across three benchmarks: COCO2014 validation set~\cite{coco}, MJHQ-30K~\cite{li2024playground}, and GenEval~\cite{geneval}.
Compared to the full attention baseline~\cite{vaswani2017attention}, prior methods employing aggressive sparsity—such as CLEAR~\cite{clear} and Neighborhood Attention (NA)~\cite{na}—exhibit significant quality degradation.
On COCO, their FID scores degrade by 8.7–13.6 points, and their image-text retrieval (IR) scores drop by an order of magnitude (from ~1.1 to 0.04–0.11).
In contrast, \attnameblock retains strong performance (FID 35.99) and dramatically boosts IR to 0.925—an 8.3$\times$ improvement over CLEAR.
\attnamecriss performs even better, closely matching the full attention baseline across all three metrics (FID 34.59, IR 1.068, CLIP-T 26.17), showing that combining local and selective long-range attention can effectively approximate dense attention.
A similar pattern holds on MJHQ-30K~\cite{li2024playground}: CLEAR and NA again lag behind, while \attnameblock stays within 1.2 FID points and 0.04 IR of full attention.
Notably, \attnamecriss achieves the best IR score (0.998) and lowest FID (20.05) among all sparse variants, matching full attention performance across all metrics.
Finally, on the reasoning-centric GenEval benchmark~\cite{geneval}, CLEAR and NA yield the lowest scores, while \attnameblock significantly improves overall performance to 0.61.
\attnamecriss further pushes the score to 0.65—on par with the full attention Flux baseline.

\textbf{Qualitative Comparison.}
Fig.~\ref{fig:visual_image} presents a visual comparison across different attention mechanisms.
Compared to CLEAR~\cite{clear} and Neighborhood Attention (NA)~\cite{na}, \attnameblock produces noticeably higher visual quality.
\attnamecriss further enhances the results, demonstrating sharper details and better global coherence through its incorporation of long-range dependencies.

\subsection{Text-to-Video Generation with HunyuanVideo}
\label{sec:result_hunyuan}

\textbf{Experimental Setup.}
We conduct text-to-video generation experiments based on HunyuanVideo~\cite{hunyuan}, which by default employs full attention implemented using FlashAttention~\cite{flashattention2}.
We replace its full attention with various sparse attention mechanisms implemented via the FlexAttention framework~\cite{flex}.
Generation quality is assessed on the VBench benchmark~\cite{vbench}.
Following prior work~\cite{sta}, we generate videos with a resolution of $768 \times 1280$ and a temporal length of 128 frames.
After VAE encoding~\cite{vae}, the resulting latent feature map has a shape of $32 \times 48 \times 80$, corresponding to a sequence length of approximately 122.9k tokens.
For our method, \attname uses a group size of $4 \times 8 \times 8$.

\textbf{Efficiency.}
Tab.~\ref{tab:speed_video} summarizes the efficiency comparison across different attention mechanisms.
Full Attention~\cite{vaswani2017attention}—the default in HunyuanVideo~\cite{hunyuan}—serves as the baseline, with each attention operation taking 0.870 seconds and the full video generation requiring 2017 seconds.
Prior methods such as Neighborhood Attention (NA)~\cite{na} and CLEAR~\cite{clear} offer modest gains, achieving 1.2$\times$ (0.721 s) and 1.3$\times$ (0.662 s) speedups in attention latency, which result in only minor improvements in end-to-end video generation time.
The concurrent work STA~\cite{sta}, tailored for video generation, achieves a 10.2$\times$ speedup (0.085 s per attention op) but restricts the farthest-token distance to just 19, limiting its receptive field.

In contrast, our proposed \attnameblock achieves the highest sparsity (94.3\%) and delivers a 15.8$\times$ speedup (0.055 s per attention op), reducing the overall video generation time from 2017 to just 546 seconds.
Despite its ``surrounding block'' design, it supports a receptive field reaching 24 tokens away—matching or surpassing existing methods with significantly lower compute.
Finally, \attnamecriss provides a balanced trade-off between speed and long-range context: by attending along both rows and columns, it covers up to 81 tokens with 60.8\% sparsity, achieving a 2.4$\times$ speedup (0.362 s) over full attention.
Overall, these results demonstrate that \attname not only outperforms existing efficient attention schemes in speed, but also offers flexible and scalable control over the receptive field—making it particularly well-suited for video diffusion tasks.

\begin{table}[t]
    \centering
    \caption{
    \textbf{Forward Speed of Sparse Attention for Video Generation.}
    All methods are based on HunyuanVideo~\cite{hunyuan} to generate 30-second videos, resulting in a sequence length of 122.9k tokens.
    Evaluation is conducted on a single A100 GPU.
    $\dagger$: Results obtained using the open-source code.
    }
    \label{tab:speed_video}
    \scalebox{0.86}{
    \begin{tabular}{c|c|ccccc} 
       \multirow{2}{*}{method}   & \multirow{2}{*}{configuration} & FLOPs & farthest & attention & inference  & inference time  \\
      & & sparsity & token& latency (s) & speedup &(sec/video)\\
      \shline
      Full Attention~\cite{hunyuan,vaswani2017attention,flashattention2}  &N/A &0\% &98  &0.870 &1.0$\times$ &2017  \\
      CLEAR~\cite{clear} &radius = 16 &89.5\% &16&0.662 &1.3$\times$&1652 \\
      NA~\cite{clear} &window = 32 & 73.3\% & 28& 0.721&1.2$\times$&1701 \\
      STA$\dagger$~\cite{sta} & window = (18, 24, 24)& 91.6\% &19&0.085 &10.2$\times$&573 \\
      \hline
      \attnameblock &b = 1 & 94.3\%&24&0.055 & 15.8$\times$ & 526\\
      \attnamecriss &N/A & 60.8\%& 81&0.362&2.4$\times$  &1102 \\
    \end{tabular}
    }
\end{table}

\textbf{Quantitative Comparison.}
Tab.~\ref{tab:vbench} presents quantitative results on the VBench~\cite{vbench}.
Our \attnameblock achieves a total score of 80.51\%, surpassing STA’s~\cite{sta} 80.46\% whiling offering a 15.8$\times$ attention speedup. This indicates that \attnameblock's locality does not substantially degrade sample quality relative to state-of-the-art sparse kernels. Notably, our \attnamecriss reaches a total score of 83.82\%, outperforming the full-attention HunyuanVideo baseline~\cite{hunyuan} (82.71\%) and all other sparse methods, while running 2.4$\times$ faster per attention call, demonstrating that complementing necessary long-range dependencies to local attentions is sufficient to maintain global coherence.

\begin{table}[t]
\centering
\caption{
\textbf{Quantitative Results of Different Attention Mechanisms on the VBench~\cite{vbench}.}
All methods are implemented using HunyuanVideo~\cite{hunyuan}.
}
\label{tab:vbench}
\begin{tabular}{c|cc|c}
method & VBench quality & VBench semantic & VBench total\\
\shline
Full Attention~\cite{hunyuan,vaswani2017attention}   &85.22\% &72.67\%         & 82.71\%  \\
CLEAR~\cite{clear}&82.81\% & 76.28\%&81.50\% \\
NA~\cite{na}& 83.01\%&77.23\%&81.85\% \\
STA~\cite{sta}       &  81.30\% &77.12\% &80.46\%   \\
\hline
\attnameblock  &81.68\% &75.82\% & 80.51\%\\
\attnamecriss & 85.35\%  & 78.01\% &83.82\% \\
\end{tabular}
\end{table}

\textbf{Qualitative Comparison.}
Fig.~\ref{fig:visual_image} presents a qualitative comparison across different attention mechanisms.
Compared to STA~\cite{sta}, our \attnameblock produces noticeably higher visual quality, while \attnamecriss further enhances the results—achieving visual fidelity comparable to full attention.

\begin{figure}[!t]
    \centering
    \setlength{\tabcolsep}{2 pt}
    \begin{tabular}{cccc}
    \includegraphics[width=0.2\linewidth]{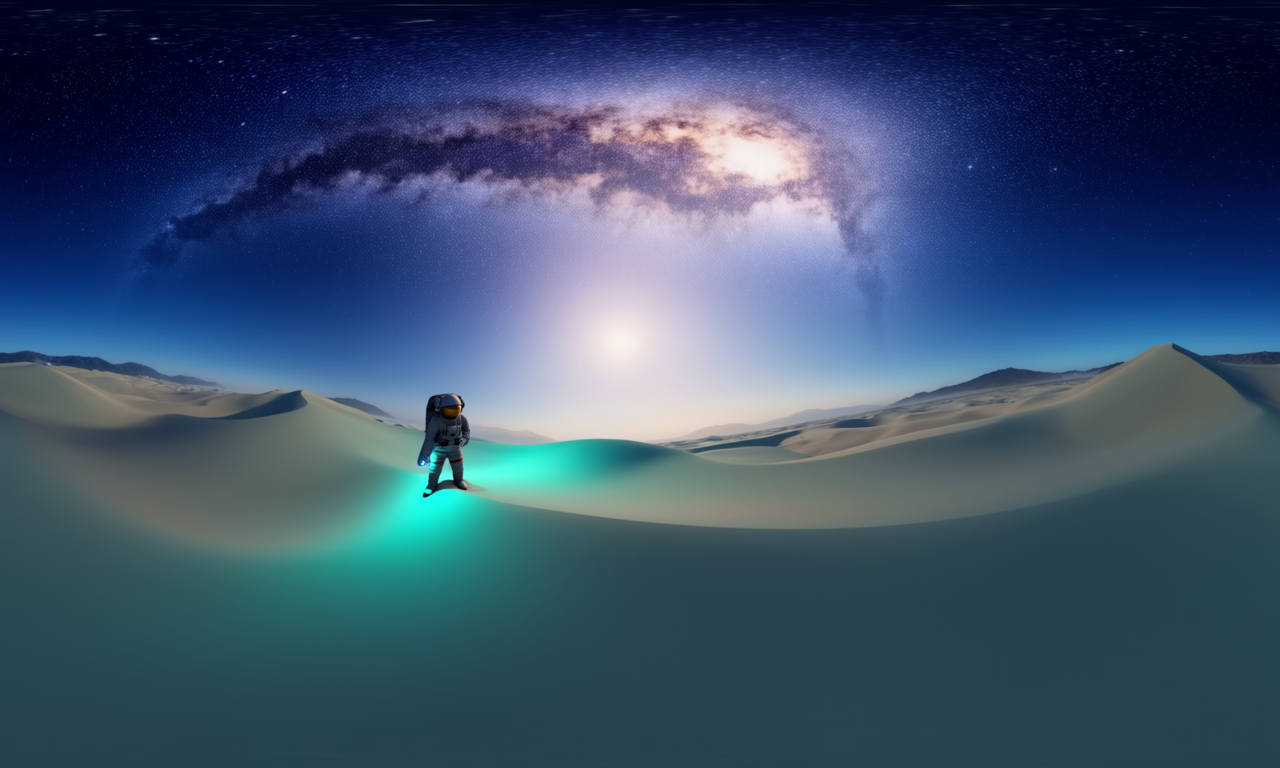} & 
        \includegraphics[width=0.2\linewidth]{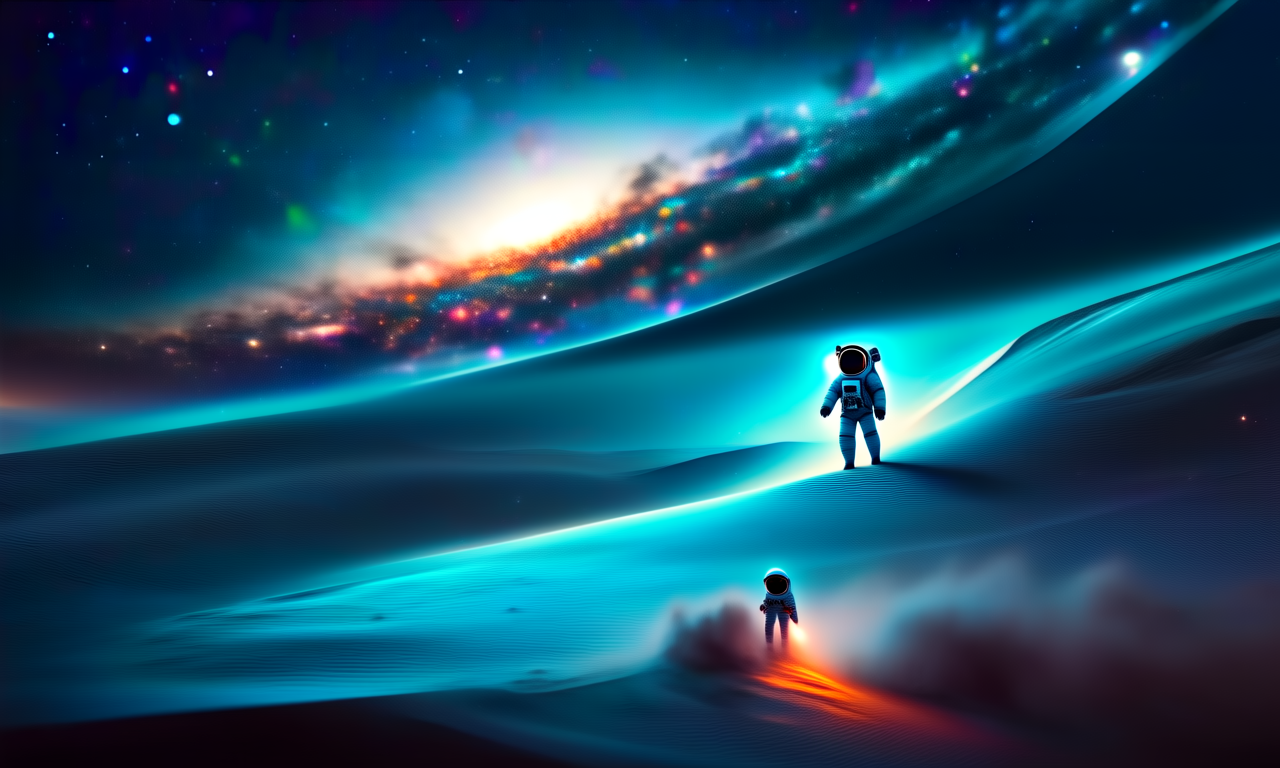} & \includegraphics[width=0.2\linewidth]{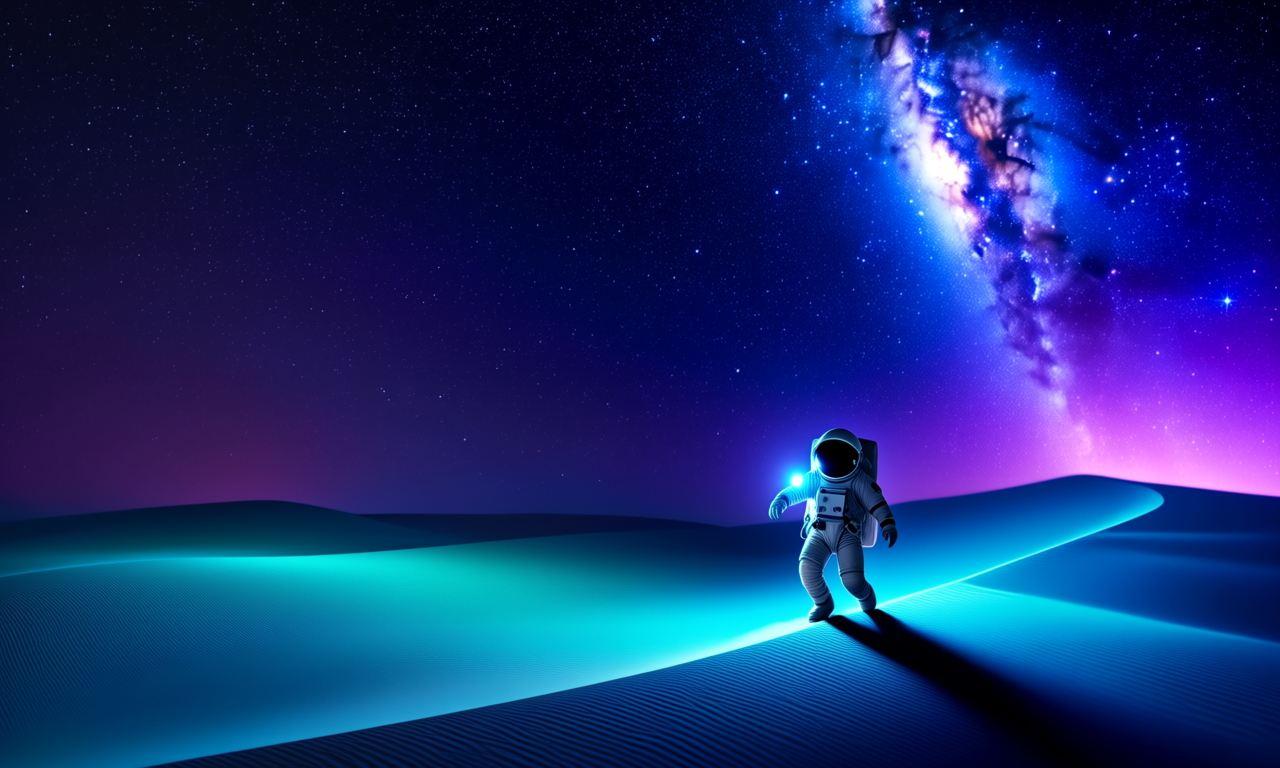} & \includegraphics[width=0.2\linewidth]{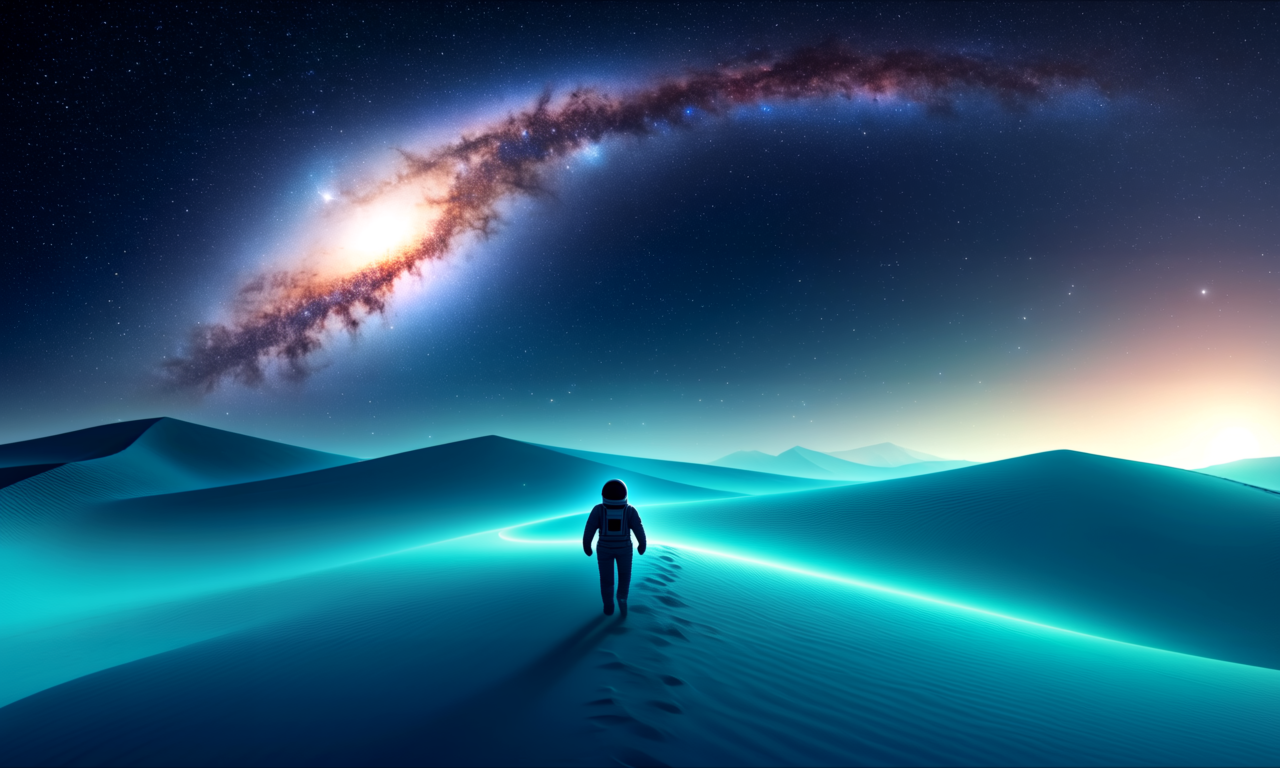}  \\
        \includegraphics[width=0.2\linewidth]{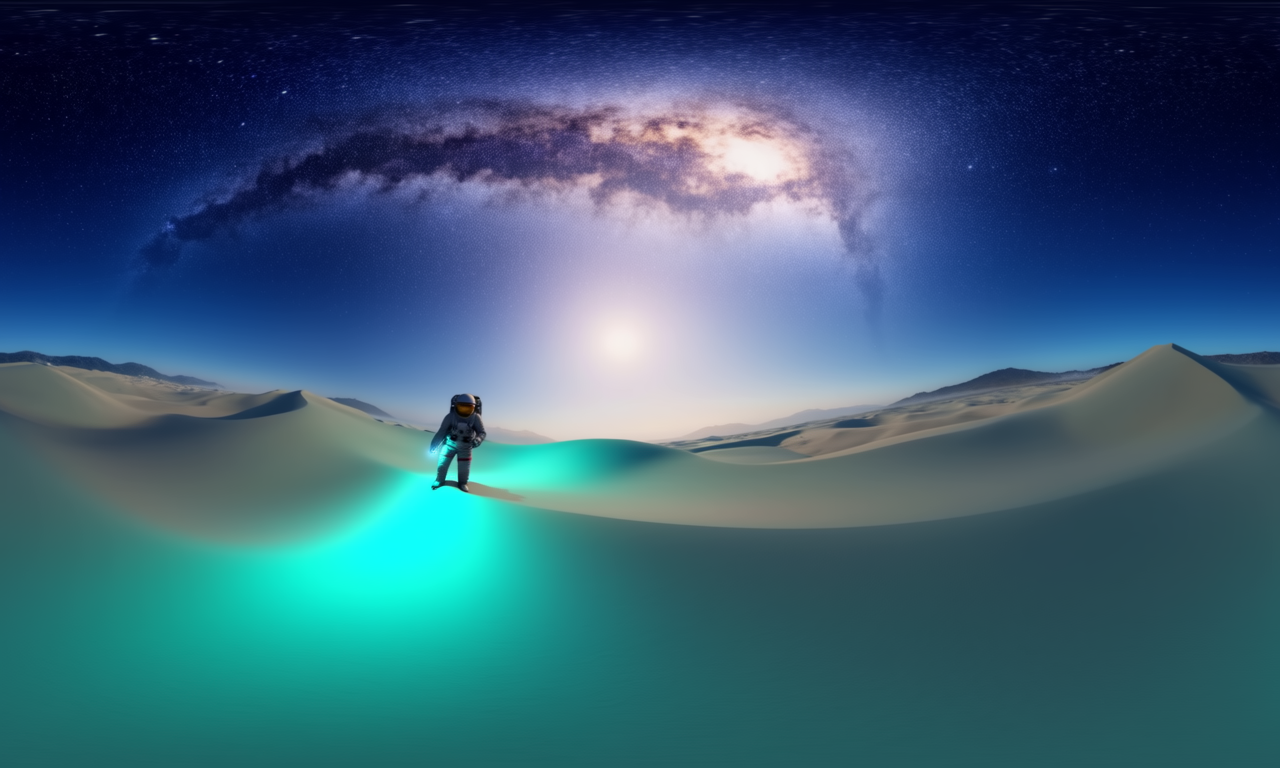} & 
        \includegraphics[width=0.2\linewidth]{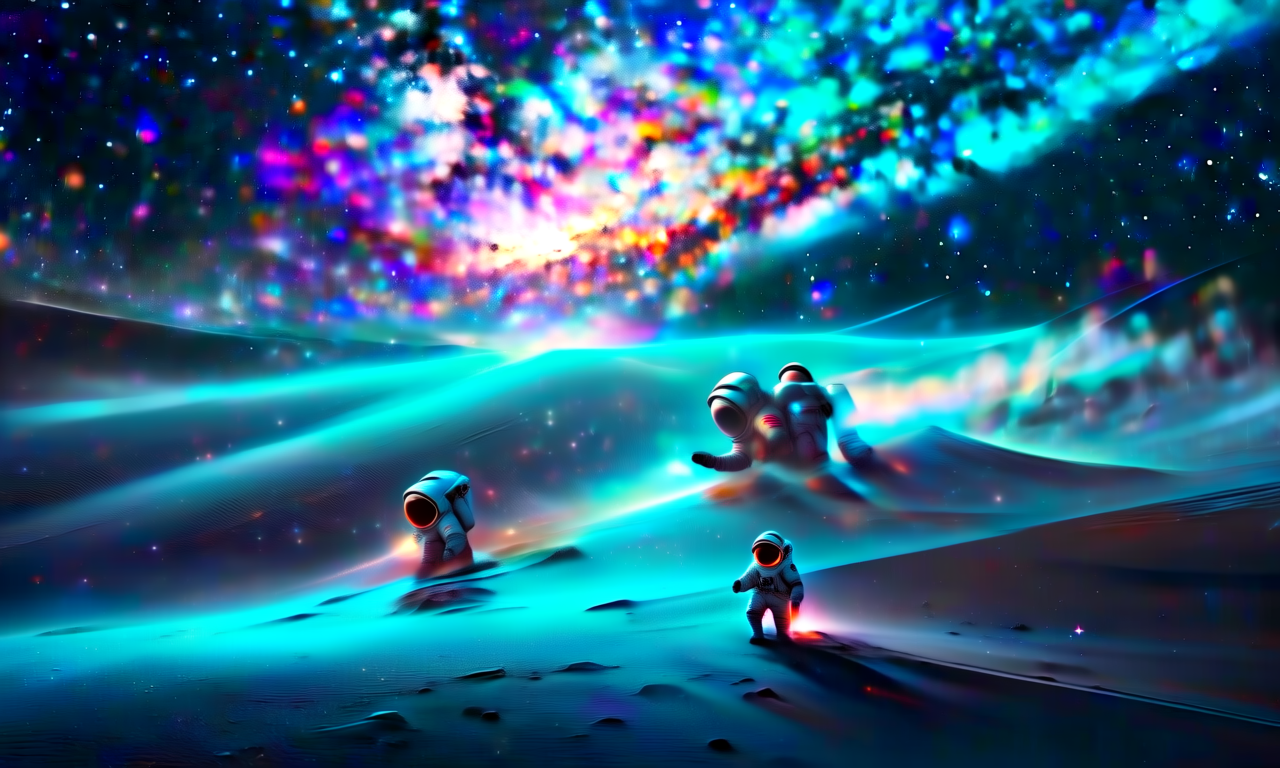} & \includegraphics[width=0.2\linewidth]{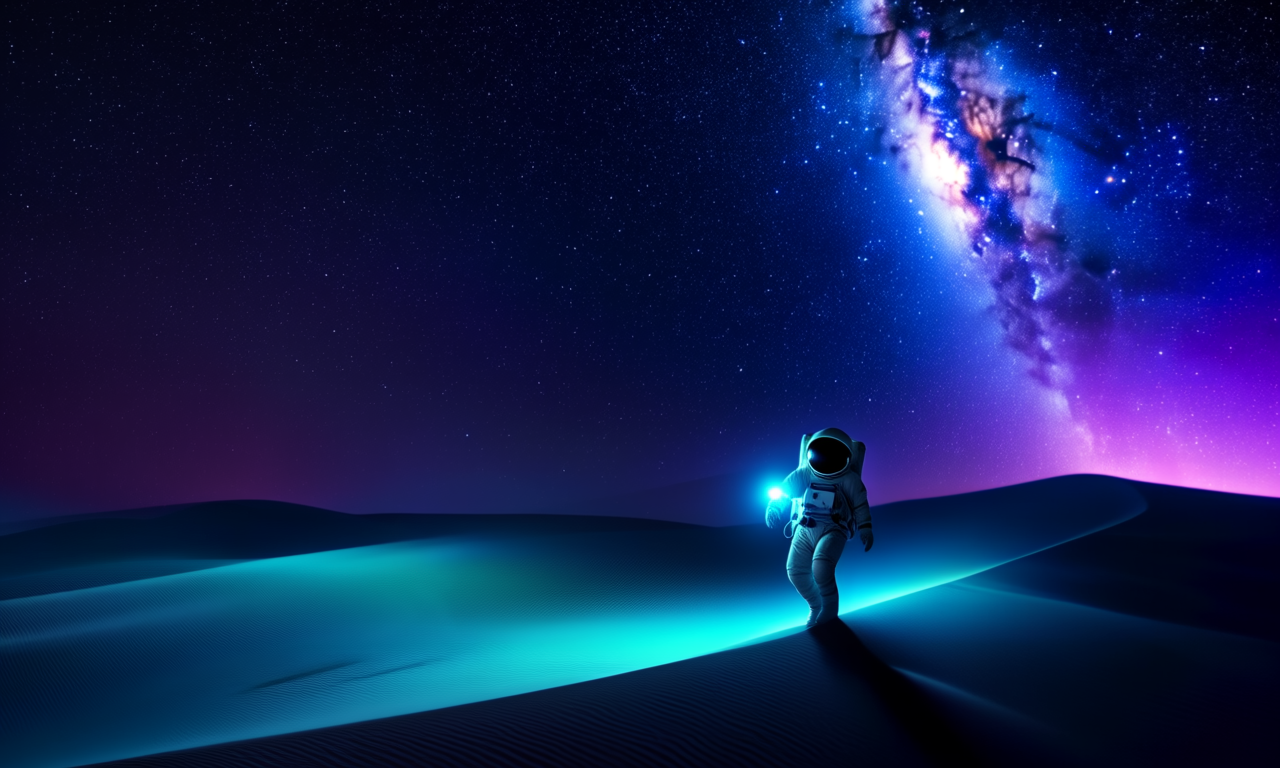} & \includegraphics[width=0.2\linewidth]{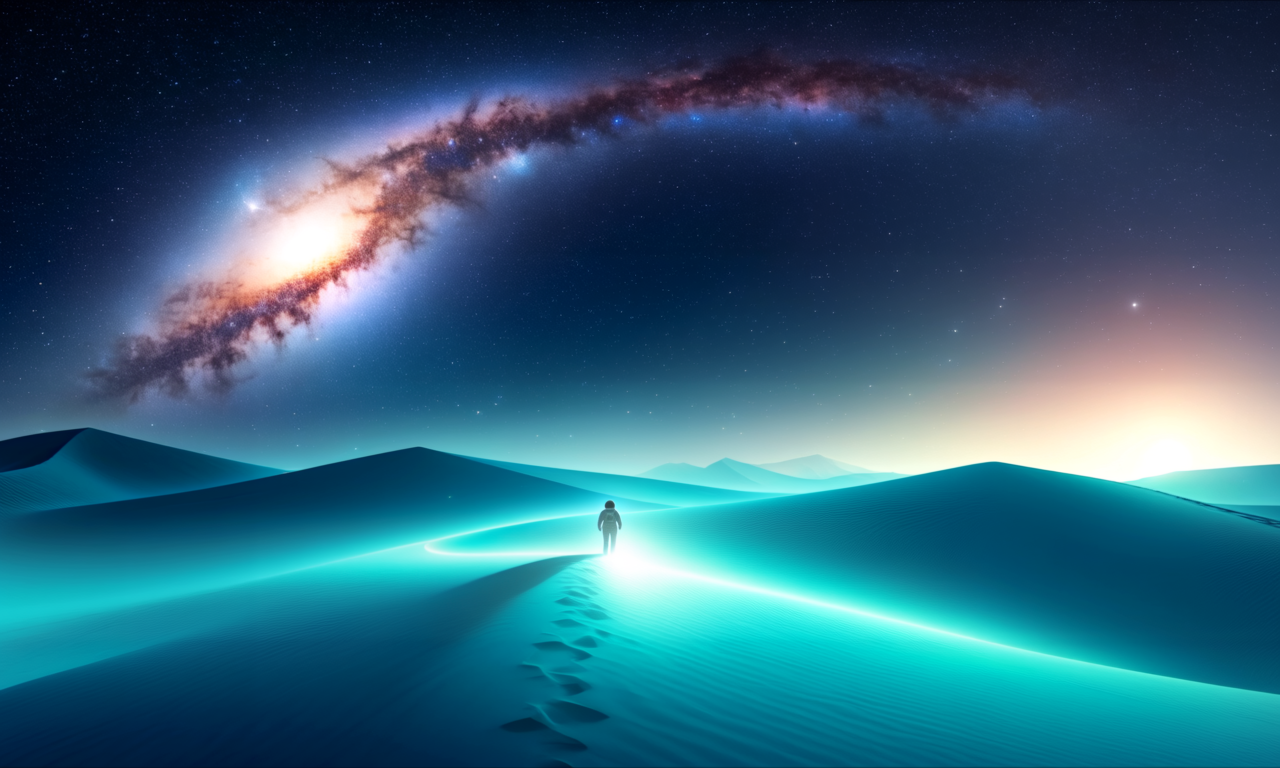}  \\
        Full Attention&STA& \attnameblock &\attnamecriss\\
    \end{tabular}
    \vspace{-2mm}
    \caption{
    \textbf{Generated Videos by HunyuanVideo~\cite{hunyuan} with Different Attention Mechanisms.}
    We compare the visual results of Full Attention~\cite{vaswani2017attention} (\ie, the original HunyuanVideo), STA~\cite{sta}, and our proposed \attnameblock and \attnamecriss.
    }
    \label{fig:visual_video}
\end{figure}

\section{Conclusion}
We introduced a training-free attention acceleration strategy that first groups input tokens and then applies structured sparse attention through surrounding and criss-cross regions.
This design achieves order-of-magnitude speedups for Diffusion Transformers while preserving critical attention patterns and essential long-range dependencies.
Extensive experiments on both image (Flux) and video (HunyuanVideo) generation demonstrate that our approach consistently outperforms prior efficient attention methods in both inference speed and generation quality, paving the way for practical deployment of high-quality diffusion models in latency-sensitive and resource-constrained scenarios.

{\small
\bibliographystyle{plainnat}
\bibliography{ref}
}
\clearpage

\appendix
\section*{Appendix}
\label{sec:supplementary}

In the appendix, we provide the following additional information:

\begin{itemize}
    \item Detailed GenEval Scores (Sec.~\ref{sec:geneval}).
    \item Ablation Study on \attname Design Choices (Sec.~\ref{sec:sup_ablation}).
    \item Additional Qualitative Visualizations (Sec.~\ref{sec:sup_vis}).
    \item Text Prompts for Generated Images and Videos (Sec.~\ref{sec:sup_prompts}).
    \item Limitation and Discussion (Sec.~\ref{sec:limitation}).
    \item Broader Impacts discussion (Sec.~\ref{sec:broader_impacts}).
    \item Dataset Licenses (Sec.~\ref{sec:dataset_license}).
\end{itemize}

\section{Detailed GenEval Scores}
\label{sec:geneval}

\begin{table}[!h]
    \centering
    \caption{
    \textbf{Quantitative Results of Different Attention Mechanisms on the GenEval~\cite{geneval}.}
    All methods are implemented using Flux.1-dev~\cite{flux}.
    }
    \label{tab:geneval}
    \scalebox{0.86}{
    \begin{tabular}{c|cccccc|c}
       method    &single obj.&two obj.&counting&colors&position&color attr. &overall$\uparrow$ \\
       \shline
       Full Attention~\cite{vaswani2017attention} &0.98&0.81&0.76&0.77&0.23&0.42&0.66 \\
      CLEAR~\cite{clear}     & 0.99&0.82&0.03&0.73&0.28&0.27&0.52\\
       NA~\cite{na}&0.98&0.82&0.12&0.74&0.25&0.31&0.54 \\
       \hline
      \attnameblock&0.98&0.84&0.48&0.75&0.26&0.37 &0.61\\
      \attnamecriss &0.98&0.84&0.62&0.76&0.30&0.39&0.65\\
    \end{tabular}
    }
\end{table}

We report detailed quantitative results on GenEval~\cite{geneval} in Tab.~\ref{tab:geneval}.
While CLEAR~\cite{clear} and Neighborhood Attention (NA)~\cite{na} demonstrate strong performance on locality-focused tasks (\eg, \textit{single/two obj.} with scores $\ge\!\!0.82$), they fail on reasoning-heavy challenges such as \textit{counting} (scoring only 0.03 and 0.12, respectively), dragging their overall scores down to the 0.52–0.54 range.
In contrast, \attnameblock achieves a notable improvement—raising the overall GenEval score to 0.61, an order-of-magnitude gain over CLEAR on the reasoning tasks.
\attnamecriss further enhances performance, leveraging long-range dependencies to reach a score of 0.65, effectively matching the full attention Flux baseline.

\section{Ablation Study}
\label{sec:sup_ablation}

\begin{table}[!h]
    \centering
    \caption{\textbf{Ablation Study.}
    We evaluate the design choices of \attname using Flux~\cite{flux}, reporting FLOPs sparsity, attention latency, and performance on the GenEval benchmark~\cite{geneval}.
    Baseline results from Full Attention~\cite{vaswani2017attention}, Neighborhood Attention (NA)~\cite{na}, and Criss-cross Attention are included for comparison.
    Our default configurations are highlighted in gray.
    }
    \scalebox{0.9}{
    \begin{tabular}{c|cc|cc|c}
       attention  &  group  size &  configuration & FLOPs sparsity& attention latency& GenEval \\
       \shline
       Full Attention~\cite{vaswani2017attention}  &  $1\times 1$ & N/A&0\%& 4.081&0.66\\
        NA~\cite{na} & $1\times 1$ & window = 32&99.42\%&0.312&0.54 \\
       Criss-cross Attention  & $1\times 1$ & N/A &99.42\%&1.640&0.58 \\
       \hline \hline
       \attnameblock &  $8\times 8$ & b = 1& 99.61\% &0.113&0.58 \\
       \attnameblock & $8\times 8$ &  b = 2& 99.22\% &0.119&0.60 \\
       \attnameblock &  $8\times 8$ & b = 3& 98.63\% &0.163&0.63 \\
       \baseline{\attnameblock} & \baseline{$16\times 16$ }& \baseline{b = 1}& \baseline{99.03\%} &\baseline{0.114}&\baseline{0.61} \\
        \attnameblock & $16\times 16$ &b = 2& 97.47\% &0.172&0.63 \\
        \attnameblock & $32\times 32$ & b = 1& 96.69\% &0.450&0.64 \\
        \hline
       \attnamecriss & $8\times 8$ & N/A & 96.71\% &0.351&0.61 \\
       \baseline{\attnamecriss} &\baseline{$16\times 16$ }& \baseline{N/A}&\baseline{93.67\%} &\baseline{0.353} & \baseline{0.65} \\
        \attnamecriss & $32\times 32$ &  N/A& 87.72\% &0.635&0.67 \\
    \end{tabular}
    }
    \label{tab:ablation}
\end{table}

In Tab.~\ref{tab:ablation}, we ablate over the group size and configuration parameters—specifically, the block size $b$ used in \attnameblock.
By default, \attnameblock uses a group size of $16 \times 16$ and $b = 1$, balancing speed and performance effectively.
Reducing the group size to $8 \times 8$ yields a slight improvement in FLOPs sparsity (from 99.03\% to 99.61\%), but attention latency remains nearly unchanged (0.114s \vs \ 0.113s), resulting in no real-world speedup. Moreover, GenEval performance drops from 0.61 to 0.58.
This suggests that overly fine group-partitioning harms GPU parallelism and memory coalescing, offsetting the theoretical FLOPs savings.
Conversely, increasing the block size (for both $8 \times 8$ and $16 \times 16$ group sizes) consistently improves GenEval performance with only modest increases in latency—indicating that attending to more neighboring blocks offers a favorable quality–efficiency tradeoff.
Similarly, for \attnamecriss, we adopt a default group size of $16 \times 16$, which achieves a balanced tradeoff between speed and generation quality.

\section{Visualizations}
\label{sec:sup_vis}

\begin{figure}
    \centering
    \includegraphics[width=\linewidth]{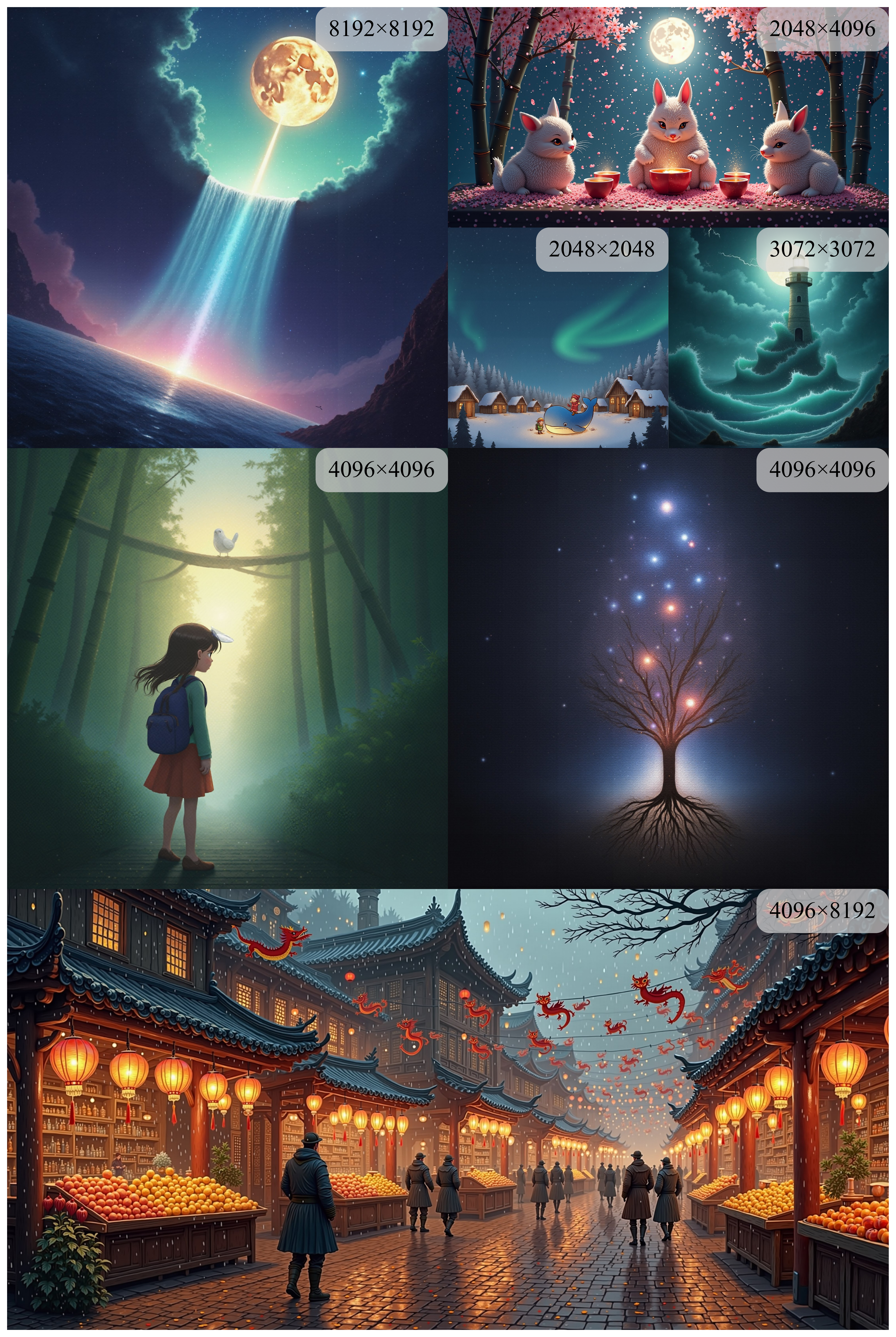}
    \caption{\textbf{Generated Images by Flux~\cite{flux} Enhanced with Our \attname Attention.}}
    \label{fig:image1}
\end{figure}

\begin{figure}
    \centering
    \includegraphics[width=\linewidth]{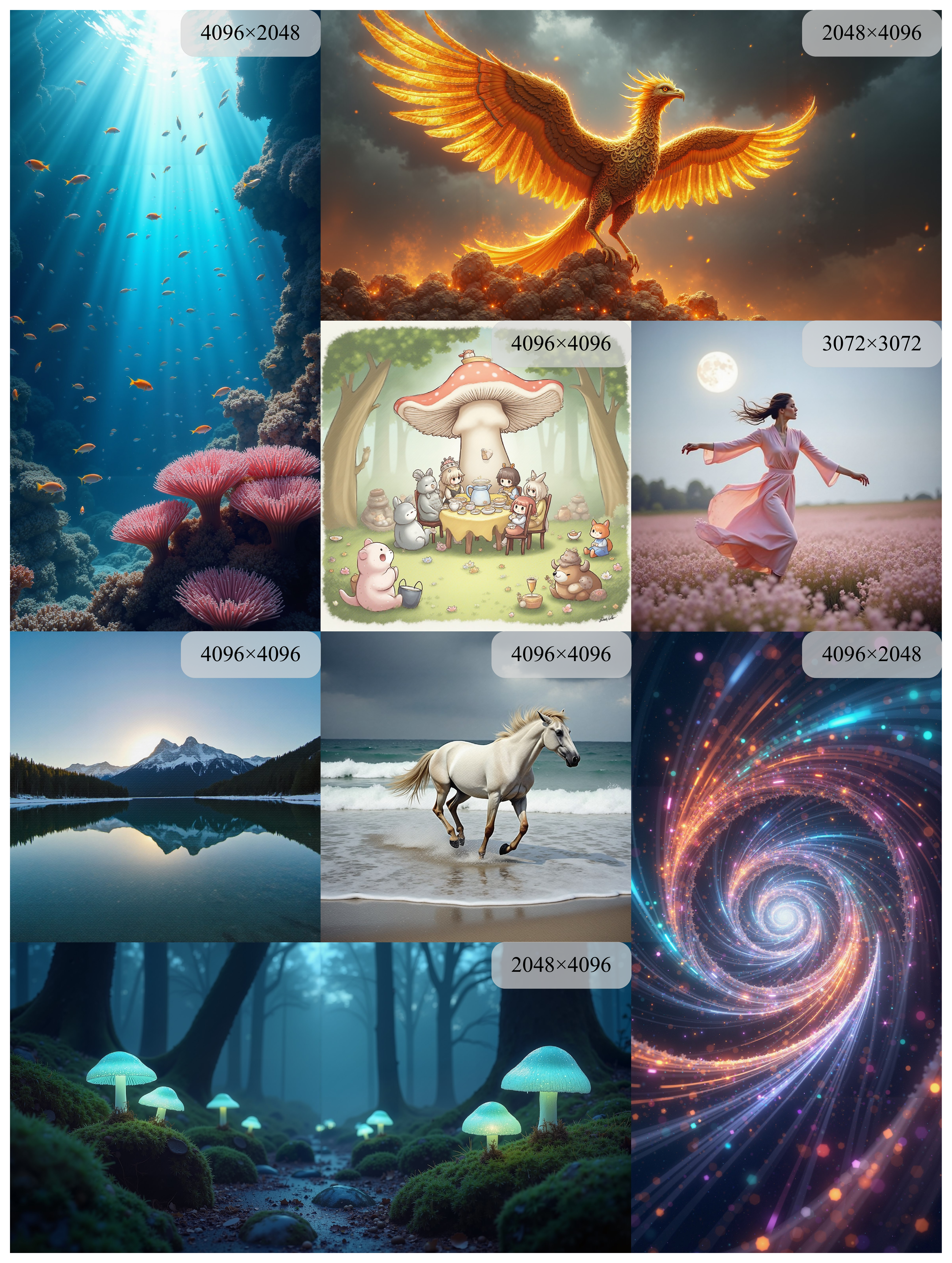}
    \caption{\textbf{Generated Images by Flux~\cite{flux} Enhanced with Our \attname Attention.}}
    \label{fig:image2}
\end{figure}

\begin{figure}
    \centering
    \includegraphics[width=\linewidth]{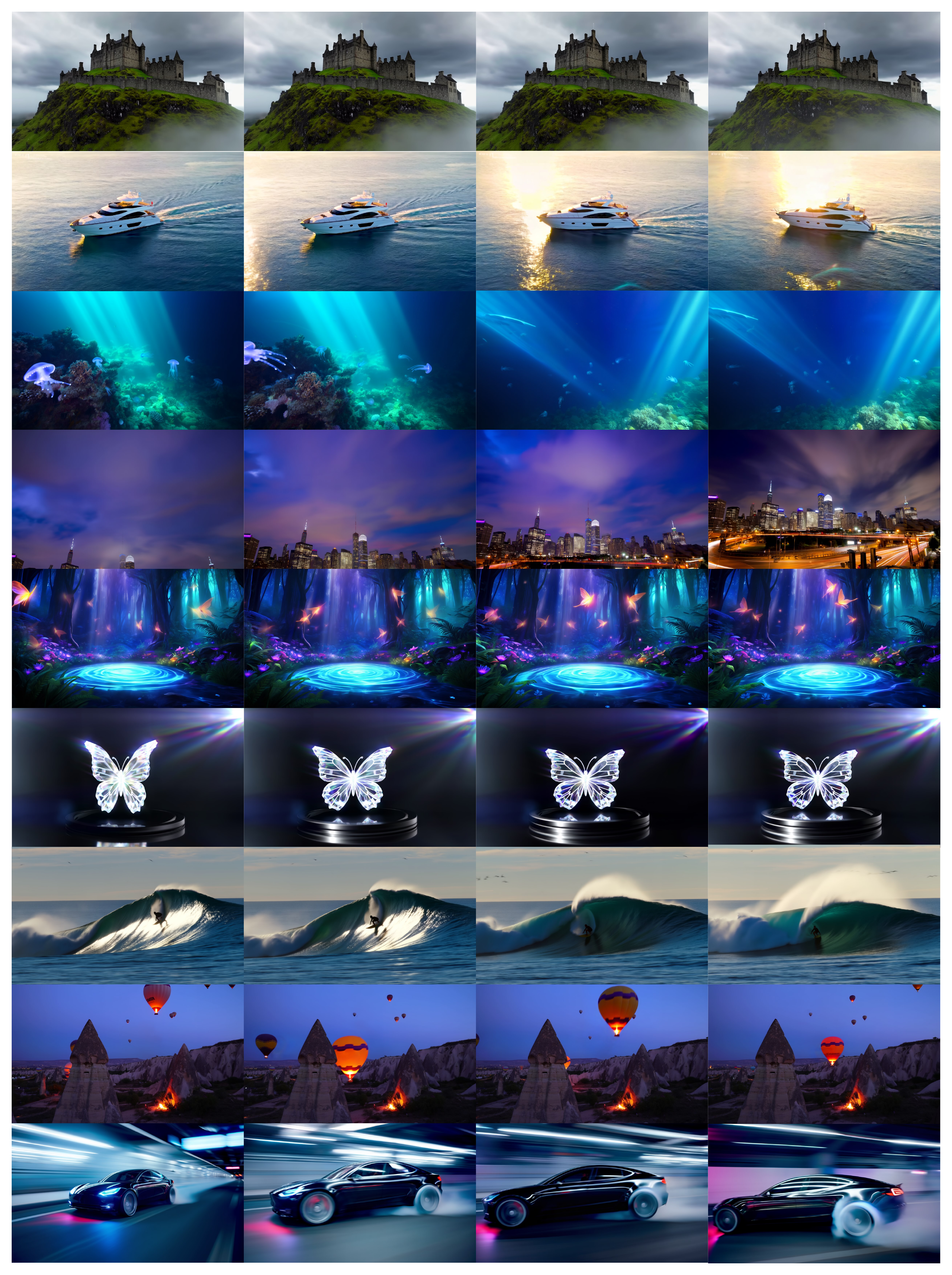}
    \caption{\textbf{Generated Videos by HunyuanVideo~\cite{hunyuan} Enhanced with Our \attname Attention.}}
    \label{fig:video}
\end{figure}

We provide additional generated images in Fig.~\ref{fig:image1} and Fig.~\ref{fig:image2}.
For video results, we show representative snapshots in Fig.~\ref{fig:video}, and include the full generated videos in the project website.
% supplementary materials.


\section{Text Prompts for Generated Images and Videos}
\label{sec:sup_prompts}
The prompts for images in Figure 1: 
\begin{itemize}
    \item Baroque ballroom reimagined as infinite fractal geometry, crystalline chandeliers in recursive cascade, holographic musicians, opalescent floor reflections, luxurious surreal elegance.
    \item Minimalist white studio background, bold 3‑D glossy letters spelling “Hello  GRAT” hovering above a reflective marble floor, soft diffused rim‑lighting, pastel gradient accents, thin golden particle streaks spiraling around the text, clean modern aesthetic.
    \item Giant glass octopus entwined around lighthouse in electric storm, lightning refracted through translucent tentacles, spray‑splashed cliffs, high‑contrast seascape.
    \item Gothic cathedral interior blooming into thorn‑rose vines, stained glass melting into liquid color, choir of spectral petals, dark romantic surrealism.
    \item Futuristic zen garden on orbiting station, raked starlight sand, floating bonsai with holographic leaves, Earthrise in panoramic window, serene sci‑fi minimalism.
    \item Crystal hummingbirds drinking from floating amethyst flowers in misty dawn garden, iridescent wings motion‑blurred, dewdrops scattering rainbow caustics, dreamy macro fantasy realism.
    
\end{itemize}

The prompts for the video in Figure 1:
\begin{itemize}
    \item  An aerial hyperlapse sweeps over a serpentine mountain road winding through autumn forests. Crimson, amber, and gold canopies blur into painterly ribbons as morning mist weaves through valleys. The camera tilts down to follow a solitary vintage car hugging the curves, engine note faintly echoing in the hush.
\end{itemize}

The prompt for images in Figure 5: 
\begin{itemize}
    \item Lava‑lit cavern arena, emerald‑robed sorcerer levitates on a magic circle, molten fireballs orbiting, crimson dragon lunging through smoke, embers swirling, ultra‑sharp depth‑of‑field, dark high‑contrast atmosphere.
\end{itemize}

The prompt for videos in Figure 6: 
\begin{itemize}
    \item Photoreal astronaut trekking across glowing turquoise desert dunes under a massive spiral galaxy sky; fine dust kicked up in low-gravity slow motion, helmet lights casting soft halos — 720 p, 24 fps, 5 s.
\end{itemize}

The prompts for images in Figure 7:
\begin{itemize}
    \item Intergalactic waterfall plummeting from fractured moon into swirling nebula ocean.
    \item Secret tea ceremony conducted by kitsune spirits beneath showering sakura petals, porcelain cups shimmering with liquid starlight, moonlit bamboo grove, serene magical ambience.
    \item Lantern‑whale festival in snowy tundra village, children riding glowing balloon cetaceans, aurora confetti sky, cozy timber cottages, whimsical winter fantasy.
    \item Dragon of living turquoise waves spiraling around obsidian lighthouse, moonlit tempest sky, lightning forks frozen midair, breathtaking elemental surrealism.
    \item Rainforest skybridge village at dawn, young girl wearing neural‑link visor guiding swarming construction nanobots weaving bamboo lattice, misty lush serenity.
    \item Tree of galaxies sprouting in deep space, branches coalescing nebulae, cosmic fruit birthing newborn stars, pilgrim ships circling roots, awe‑inspiring vastness.
    \item Medieval market square during lantern festival, paper dragons swirling on invisible strings, stalls glowing with alchemical fruits, cobblestone slick from gentle rain, painterly warm atmosphere.
\end{itemize}

The prompts for images in Figure 8:
\begin{itemize}
\item Underwater coral reef teeming with neon fish, shafts of sunlight piercing blue waters, dynamic coral formations, photorealistic clarity.
\item A golden mechanical phoenix rising from molten gears against a stormy sky, intricate filigree wings glinting with fiery embers.
\item Whimsical tea party in a giant mushroom grove with tiny anthropomorphic animals, pastel colors, storybook charm.
\item Ethereal dancer in flowing silk robes spinning in a blossom-filled meadow under a full moon, motion blur, soft pastel palette.
\item Snow-capped mountain reflected in a crystal-clear alpine lake at sunrise, mirror symmetry, serene tranquility.
\item Majestic white stallion galloping across a stormy beach with crashing waves, dynamic movement, dramatic sky.
\item Abstract fractal vortex swirling with iridescent ribbons of light, infinite tunnel effect, high detail, cosmic color spectrum.
\item A surreal luminescent forest glade at twilight, bioluminescent mushrooms casting soft glows on mossy stones, swirling mist, otherworldly ambiance.
\end{itemize}

The prompts for videos in Figure 9: 
\begin{itemize}
    \item A medieval castle perched atop a mist-shrouded hill under brooding clouds. The camera ascends in a smooth drone shot, revealing moss-covered battlements and fluttering pennants. Occasional shafts of sunlight pierce the gloom, illuminating arrow slits and creeping ivy, while distant horns echo across the fog-laden valley.
    \item A sleek white yacht gliding across a crystal-blue sea at sunset, camera circles the vessel as golden light sparkles on gentle waves, slight lens distortion.
    \item An underwater dreamscape of neon-glowing jellyfish drifting through a cathedral-like coral reef. Shafts of turquoise solar light pierce the sapphire depths, illuminating swaying anemones. The camera tracks laterally at midwater, capturing the creatures’ delicate tendrils and pulsing bioluminescence against the hushed chorus of filtered waves.
    \item A time-lapse of a global metropolis at dusk: skyscraper lights blink awake one by one under a watercolor sky. Clouds streak past in accelerated motion while the camera cranes upward from bustling street intersections to soaring heights, revealing a tapestry of neon and chrome. Subtle ambient city hum crescendos.
    \item A fantasy forest glade alive with bioluminescent mushrooms and ferns pulsing in time with an unseen etheric heartbeat. Firefly-like sprites flit through shafts of moonlight. The camera slowly orbits a crystalline pool at the center, its surface rippling with phosphorescent ripples, evoking a dreamlike pulse of magic.
    \item A CGI crystalline butterfly emerges from a luminescent cocoon in slow unfolding. Its translucent wings catch prismatic spotlights, scattering rainbow glints across a reflective black pedestal. The camera executes a tight half-circle, revealing intricate geometric veins and pulsing inner glow, set to a soft ambient chime.
    \item A high-contrast silhouette of a lone surfer catching a wave at dawn; camera plunges beneath the rolling crest, capturing sunlit ripples refracted through translucent blue water. Bursts of foam and lens flare accent every curve, while distant seabirds cry over a pastel horizon.
    \item A twilight shot of lantern-lit hot-air balloons ascending over Cappadocian valleys; whimsical craft drift past tufa pillars as camera tracks upward. Embers from onboard burners glow against the fading sky, while distant festival music hints at celebration in the cool evening breeze.
    \item A futuristic concept car drifting through neon-lit urban tunnels; glossy chassis reflecting LED strips overhead. The camera follows from a low angle, catching tire smoke and subtle underbody glow, as an electronic bass beat accelerates in sync with the vehicle’s ghostly motion.
\end{itemize}

\section{Limitation and Discussion}
\label{sec:limitation}
While \attname achieves significant speedups and preserves generation quality across both image and video domains, it has a few limitations. The grouping and structured attention patterns—though inspired by learned attention behaviors—are manually defined and fixed at inference time, which may limit adaptability across diverse inputs or tasks.
Future work could explore adaptive or learnable grouping strategies and data-dependent sparsity patterns to enhance flexibility and extend applicability to a broader range of computer vision tasks.

\section{Broader Impacts}
\label{sec:broader_impacts}
This work contributes to improving the accessibility, scalability, and environmental sustainability of high-quality generative models. By introducing a training-free acceleration strategy, \attname reduces the inference cost of Diffusion Transformers without sacrificing sample quality—making state-of-the-art image and video generation feasible on more modest hardware. This has the potential to democratize generative AI, enabling wider use in education, creativity, accessibility tools, and small-scale research. Additionally, the reduced computational footprint contributes to energy efficiency, addressing growing concerns about the carbon impact of large AI models. However, as with all generative technologies, the broader deployment of such systems also raises risks related to misuse, including misinformation, deepfakes, or biased content generation. We encourage developers and researchers to adopt responsible deployment practices, including model transparency, usage guidelines, and robust content moderation, to mitigate these potential harms.

\section{Dataset Licenses}
\label{sec:dataset_license}

The proposed \attname is a training-free method. Therefore, we provide the evaluation dataset licenses below:

\textbf{COCO2014}: Following prior work~\cite{clear}, we randomly sample 5,000 images along with their prompts from the COCO 2014 validation set~\cite{coco} for text-to-image evaluation.

License: \href{https://cocodataset.org/#termsofuse}{https://cocodataset.org/termsofuse} 

URL: \href{https://cocodataset.org/#home}{https://cocodataset.org} 

\textbf{MJHQ-30K}: 
MJHQ-30K~\cite{li2024playground} is used for automatic evaluation of a model’s aesthetic quality. We randomly sample 5,000 images along with their prompts for text-to-image evaluation.

License: \href{https://github.com/zsxkib/playground-v2-1024px-aesthetic/blob/master/LICENSE}{https://github.com/zsxkib/playground-v2-1024px-aesthetic/blob/master/LICENSE}

URL: \href{https://github.com/zsxkib/playground-v2-1024px-aesthetic}{https://github.com/zsxkib/playground-v2-1024px-aesthetic}

\textbf{GenEval}: The GenEval dataset~\cite{geneval} consists of 553 prompts with four images generated per prompt. The generated images are evaluated in terms of various criteria (such as single object, two object, counting, \etc).

License: \href{https://github.com/djghosh13/geneval/blob/main/LICENSE}{https://github.com/djghosh13/geneval/blob/main/LICENSE}

URL: \href{https://github.com/djghosh13/geneval}{https://github.com/djghosh13/geneval}

\textbf{VBench}: VBench~\cite{vbench} designs a comprehensive and hierarchical Evaluation Dimension Suite to decompose ``video generation quality'' into multiple well-defined dimensions to facilitate fine-grained and objective evaluation. It consists of 946 prompts.

License: \href{https://github.com/Vchitect/VBench/blob/master/LICENSE}{https://github.com/Vchitect/VBench/blob/master/LICENSE}

URL: \href{https://github.com/Vchitect/VBench}{https://github.com/Vchitect/VBench}

\end{document}